# High Dimensional Semiparametric Scale-Invariant Principal Component Analysis

Fang Han* and Han Liu†

February 18, 2014

## Abstract

We propose a new high dimensional semiparametric principal component analysis (PCA) method, named Copula Component Analysis (COCA). The semiparametric model assumes that, after unspecified marginally monotone transformations, the distributions are multivariate Gaussian. COCA improves upon PCA and sparse PCA in three aspects: (i) It is robust to modeling assumptions; (ii) It is robust to outliers and data contamination; (iii) It is scale-invariant and yields more interpretable results. We prove that the COCA estimators obtain fast estimation rates and are feature selection consistent when the dimension is nearly exponentially large relative to the sample size. Careful experiments confirm that COCA outperforms sparse PCA on both synthetic and real-world datasets.

## 1 Introduction

In this paper we propose a new Principal Component Analysis (PCA), named Copula Component Analysis (COCA), based on a semiparametric model for analyzing high dimensional non-Gaussian data. The semiparametric model assumes that, after marginal-wise unspecified strictly increasing transformations, the data are Gaussian distributed. This model is proposed by Liu et al. (2009) and a rank-based estimator for inferring graphical models is proposed by Liu et al. (2012). We generalize their results to estimate the leading eigenvectors of the correlation and covariance matrices. New estimation methods and their theoretical and empirical performances are provided.

Let $\boldsymbol{X} \in \mathbb{R}^d$ be the random vector with interest to us. PCA aims at recovering the top $m$ leading eigenvectors $\boldsymbol{u}_1, \ldots, \boldsymbol{u}_m$ of $\boldsymbol{\Sigma} := \text{Cov}(\boldsymbol{X})$. In practice, $\boldsymbol{\Sigma}$ is unknown and is replaced by the sample covariance matrix $\mathbf{S}$ using $n$ independent realizations of $\boldsymbol{X}$. For fixed $d$, PCA always achieves a consistent estimator and its asymptotic efficiency property is well addressed (Anderson, 1958). However, under a doubly asymptotic framework in which both the sample size $n$ and dimensionality $d$ can increase (with possibly $d > n$), Johnstone and Lu (2009) showed that the leading eigenvector of $\mathbf{S}$ cannot converge to $\boldsymbol{u}_1 = (u_{11}, \ldots, u_{1d})^T$. A common remedy is to assume that $s := \text{card}(\{j : u_{1j} \neq 0\}) < n$. Different sparse PCA algorithms are being developed to exploit this sparsity structure and we refer to, Yuan and Zhang (2013), Ma (2013), and Vu and Lei (2012), among others.

---

*Department of Biostatistics, Johns Hopkins University, Baltimore, MD 21205, USA; e-mail: fhan@jhsph.edu

†Department of Operations Research and Financial Engineering, Princeton University, Princeton, NJ 08544, USA; e-mail: hanliu@princeton.edu

There are several drawbacks of PCA and sparse PCA: (i) Data are assumed to be Gaussian or sub-Gaussian distributed such that a fast convergence rate can be obtained; (ii) They are not scale-invariant, i.e., changing the measurement scale of variables makes the estimates different (Borgognone et al., 2001); (iii) They are not robust to data contaminations (outliers, for example). To address these concerns, we propose a high dimensional semiparametric scale-invariant principal component analysis method, named COpula Component Analysis (COCA), based on the nonparanormal family. Here we say that $\boldsymbol{X} = (X_1, \ldots, X_d)^T$ is nonparanormally distributed if there exists a set of univariate strictly increasing functions $f = \{f_j\}_{j=1}^d$ such that $(f_1(X_1), \ldots, f_d(X_d))^T \sim N_d(\mathbf{0}, \boldsymbol{\Sigma}^0)$. By treating the monotone transformation functions $\{f_j\}_{j=1}^d$ as a type of data contamination, COCA aims at recovering the leading eigenvectors of the *latent correlation matrix* $\boldsymbol{\Sigma}^0$.

Compared with PCA and sparse PCA, COCA is scale-invariant and its estimating procedure is adaptive over the whole nonparanormal family. The nonparanormal family contains and is much larger distribution family than the Gaussian. By exploiting a rank-based regularized procedure for parameter estimation, the COCA estimator is not only robust to modeling and data contaminations, but can be consistent even when the dimensionality is nearly exponentially large relative to the sample size.

In this paper, to complete the story, a scale variant PCA method, named Copula PCA, is also proposed. Copula PCA estimates the leading eigenvector of the latent covariance matrix $\boldsymbol{\Sigma}$ (detailed definition provided in Section 2.2). To estimate $\boldsymbol{\Sigma}$, instead of $\boldsymbol{\Sigma}^0$, in a fast rate, we prove that extra conditions are required on the transformation functions.

Liu et al. (2012) proposed a procedure called the nonparanormal SKEPTIC to estimate the graphical model via exploiting the nonparanormal distribution to model the data and rank based methods for estimation. COCA is different from the nonparanormal SKEPTIC in three aspects: (i) Their focus is on graph estimation, in contrast, this paper focuses on PCA and propose new estimation methods with thorough theoretical analysis provided; (ii) We provide a second step projection to make the estimated rank-based correlation and covariance matrices positive semidefinite, and prove that the same parametric rate can be preserved; (iii) Unlike the previous analysis, this paper provides extra conditions on the transformation functions to guarantee the fast rates of convergence for Copula PCA, and we discuss the advantages of COCA over Copula PCA.

The rest of the paper is organized as follows. In the next section, we briefly discuss the statistical model of the scale-invariant PCA and review the nonparanormal model and rank-based estimators shown in Liu et al. (2009, 2012). In Section 3, we present the model of COCA and introduce the corresponding estimators and algorithms. We provide a theoretical analysis of COCA estimators in Section 4. In section 5, we employ COCA on both synthetic and real-world data to show its empirical usefulness. Some of the results in this paper were first stated without proofs in a conference version (Han and Liu, 2012).

# 2 Background

We start with notations: Let $\mathbf{M} = [\mathbf{M}_{jk}] \in \mathbb{R}^{d \times d}$ and $\boldsymbol{v} = (v_1, ..., v_d)^T \in \mathbb{R}^d$. Let $\boldsymbol{v}$'s subvector with entries indexed by $I$ be denoted by $\boldsymbol{v}_I$. Let $\mathbf{M}$'s submatrix with rows indexed by $I$ and columns indexed by $J$ be denoted by $\mathbf{M}_{IJ}$. Let $\mathbf{M}_{I*}$ and $\mathbf{M}_{*J}$ be the submatrix of $\mathbf{M}$ with rows in $I$, and the submatrix of $\mathbf{M}$ with columns in $J$. For $0 < q < \infty$, we define the $\ell_q$ and $\ell_\infty$ vector norms as $\|\boldsymbol{v}\|_q := (\sum_{i=1}^d |v_i|^q)^{1/q}$ and $\|\boldsymbol{v}\|_\infty := \max_{1 \le i \le d} |v_i|$, and we define $\|\boldsymbol{v}\|_0 := \text{card}(\text{supp}(\boldsymbol{v}))$. Here card$(\cdot)$ represents the cardinality and supp$(\boldsymbol{v}) := \{j : v_j \ne 0\}$. We define the matrix $\ell_{\max}$ norm as the elementwise maximum value: $\|\mathbf{M}\|_{\max} := \max\{|\mathbf{M}_{ij}|\}$. We define Tr$(\mathbf{M})$ to be the trace of $\mathbf{M}$. Let $\Lambda_j(\mathbf{M})$ be the $j$-th largest eigenvalue of $\mathbf{M}$. In particular, $\Lambda_{\min}(\mathbf{M}) := \Lambda_d(\mathbf{M})$ and $\Lambda_{\max}(\mathbf{M}) := \Lambda_1(\mathbf{M})$ are the smallest and largest eigenvalues of $\mathbf{M}$. The vectorized matrix of $\mathbf{M}$, denoted by vec$(\mathbf{M})$, is defined as vec$(\mathbf{M}) := (\mathbf{M}_{*1}^T, \ldots, \mathbf{M}_{*d}^T)^T$. Let $\mathbb{S}^{d-1} := \{\boldsymbol{v} \in \mathbb{R}^d : \|\boldsymbol{v}\|_2 = 1\}$ be the $d$-dimensional $\ell_2$ sphere. For any two vectors $\boldsymbol{a}, \boldsymbol{b} \in \mathbb{R}^d$ and any two squared matrices $\mathbf{A}, \mathbf{B} \in \mathbb{R}^{d \times d}$, denote the inner product of $\boldsymbol{a}$ and $\boldsymbol{b}$, $\mathbf{A}$ and $\mathbf{B}$ by $\langle \boldsymbol{a}, \boldsymbol{b} \rangle := \boldsymbol{a}^T \boldsymbol{b}$ and $\langle \mathbf{A}, \mathbf{B} \rangle := \text{Tr}(\mathbf{A}^T \mathbf{B})$. Let diag$(\mathbf{M}) := (\mathbf{M}_{11}, \mathbf{M}_{22}, \ldots, \mathbf{M}_{dd})^T$. we denote sign$(\boldsymbol{a}) := (\text{sign}(a_1), \ldots, \text{sign}(a_d))^T$, where sign$(x) = x/|x|$ with the convention $0/0 = 0$.

## 2.1 The Models of PCA and Scale-Invariant PCA

PCA is not scale-invariant, meaning that variables measured in different scales will result in different estimators (Flury, 1997). To attack this problem, PCA conducted on the sample correlation matrix $\mathbf{S}^0$ instead of the sample covariance matrix $\mathbf{S}$ is commonly used. We call the procedure of conducting PCA on $\mathbf{S}^0$ the scale-invariant PCA. It is realized that a large portion of works claiming doing PCA are actually doing the scale-invariant PCA (Borgognone et al., 2001), and the theoretical performance of the scale-invariant PCA in low dimensionals has been studied (Konishi, 1979; Nagao, 1988). It is under debate whether PCA or the scale-invariant PCA are preferred in different circumstances and we refer to Chatfield and Collins (1980), Flury (1997), and Johnson and Wichern (2007) for more discussions on it.

Let $\boldsymbol{\Sigma}^0$ and $\boldsymbol{\Sigma}$ be the correlation and covariance matrices of a random vector $\boldsymbol{X} \in \mathbb{R}^d$. Let $\omega_1 \ge \omega_2 \ge \ldots \ge \omega_d > 0$ and $\lambda_1 \ge \lambda_2 \ge \ldots \ge \lambda_d > 0$ be the eigenvalues of $\boldsymbol{\Sigma}$ and $\boldsymbol{\Sigma}^0$. Let $\boldsymbol{u}_1, \ldots, \boldsymbol{u}_d$ and $\boldsymbol{\theta}_1, \ldots, \boldsymbol{\theta}_d$ be the corresponding eigenvectors. The next proposition claims that the estimators $\{\widehat{\boldsymbol{u}}_1, \ldots, \widehat{\boldsymbol{u}}_d\}$ and $\{\widehat{\boldsymbol{\theta}}_1, \ldots, \widehat{\boldsymbol{\theta}}_d\}$, which are the eigenvectors of the sample covariance and correlation matrices $\mathbf{S}$ and $\mathbf{S}^0$, are the Maximum Likelihood Estimators (MLEs) of $\{\boldsymbol{u}_1, \ldots, \boldsymbol{u}_d\}$ and $\{\boldsymbol{\theta}_1, \ldots, \boldsymbol{\theta}_d\}$:

**Proposition 2.1** (Anderson (1958)). *Let $\boldsymbol{X} \sim N_d(\boldsymbol{\mu}, \boldsymbol{\Sigma})$ and $\boldsymbol{\Sigma}^0$ be the correlation matrix of $\boldsymbol{X}$. Let $\boldsymbol{x}_1 \ldots \boldsymbol{x}_n$ be $n$ independent realizations of $\boldsymbol{X}$. Then the estimators of PCA, $\{\widehat{\boldsymbol{u}}_1, \ldots, \widehat{\boldsymbol{u}}_d\}$, and the estimators of the scale-invariant PCA, $\{\widehat{\boldsymbol{\theta}}_1, \ldots, \widehat{\boldsymbol{\theta}}_d\}$, are the MLEs of $\{\boldsymbol{u}_1, \ldots, \boldsymbol{u}_d\}$ and $\{\boldsymbol{\theta}_1, \ldots, \boldsymbol{\theta}_d\}$.*

The scale-invariant PCA is a safe procedure for dimension reduction when variables are measured in different scales. In this paper we further show that under a more general nonparanormal (or Gaussian copula) model, the scale-invariant PCA will pose less conditions than PCA to make the estimators achieve good theoretical performance.

## 2.2 The Nonparanormal Distribution

We first introduce the two definitions of the nonparanormal distribution separately shown in Liu et al. (2009) and Liu et al. (2012). These two definitions will be used to define the models of COCA and Copula PCA in the next section.

**Definition 2.1** (Liu et al. (2009)). A random vector $\boldsymbol{X} = (X_1, ..., X_d)^T$ with means $\boldsymbol{\mu} = (\mu_1, \ldots, \mu_d)^T$ and standard deviations $\{\sigma_1, \ldots, \sigma_d\}$ is said to follow a margin-preserved nonparanormal distribution $MNPN_d(\boldsymbol{\mu}, \boldsymbol{\Sigma}, f)$ if and only if there exists a set of strictly increasing univariate functions $f = \{f_j\}_{j=1}^d$ such that:
$$f(\boldsymbol{X}) = (f_1(X_1), ..., f_d(X_d))^T \sim N_d(\boldsymbol{\mu}, \boldsymbol{\Sigma}),$$
where $\text{diag}(\boldsymbol{\Sigma}) = (\sigma_1^2, \ldots, \sigma_d^2)^T$. We call $\boldsymbol{\Sigma}$ the *latent covariance matrix*.

**Definition 2.2** (Liu et al. (2012)). Let $f^0 = \{f_j^0\}_{j=1}^d$ be a set of strictly increasing univariate functions. We say that a $d$ dimensional random vector $\boldsymbol{X} = (X_1, \ldots, X_d)^T$ follows a nonparanormal distribution $NPN_d(\boldsymbol{\Sigma}^0, f^0)$, if
$$f^0(\boldsymbol{X}) := (f_1^0(X_1), \ldots, f_d^0(X_d))^T \sim N_d(\boldsymbol{0}, \boldsymbol{\Sigma}^0),$$
where $\text{diag}(\boldsymbol{\Sigma}^0) = \boldsymbol{1}$. We call $\boldsymbol{\Sigma}^0$ the *latent correlation matrix*.

We have the following lemma, which proves that the two definitions of the nonparanormal are equivalent.

**Lemma 2.2.** A random vector $\boldsymbol{X} \sim NPN_d(\boldsymbol{\Sigma}^0, f^0)$ if and only if there exists $\boldsymbol{\mu} = (\mu_1, \ldots, \mu_d)^T$, $\boldsymbol{\Sigma} = [\boldsymbol{\Sigma}_{jk}] \in \mathbb{R}^{d \times d}$ with
$$\mathbb{E}(X_j) = \mu_j, \ \text{Var}(X_j) = \boldsymbol{\Sigma}_{jj} \quad \text{and} \quad \boldsymbol{\Sigma}_{jk}^0 = \frac{\boldsymbol{\Sigma}_{jk}}{\sqrt{\boldsymbol{\Sigma}_{jj} \cdot \boldsymbol{\Sigma}_{kk}}},$$
and a set of strictly increasing univariate functions $f = \{f_j\}_{j=1}^d$ such that $\boldsymbol{X} \sim MNPN_d(\boldsymbol{\mu}, \boldsymbol{\Sigma}, f)$.

*Proof.* Using the connection that $f_j(\cdot) = \mu_j + \sigma_j f_j^0(\cdot)$, for $j \in \{1, 2, \ldots, d\}$. □

Liu et al. (2009) proved that the nonparanormal family is equivalent to the continuous Gaussian copula family (Klaassen and Wellner, 1997). Definition 2.2 is more appealing because it emphasizes the correlation and hence matches the spirit of the copula. However, Definition 2.1 enjoys notational simplicity in analyzing the nonparanormal based linear discriminant analysis and scale-variant PCA methods.

Here we note that in Definition 2.2, the model is identifiable. Moreover, the parameters $\boldsymbol{\mu}$ and $\boldsymbol{\Sigma}$ in the latent Gaussian random vector $f(\boldsymbol{X}) \sim N_d(\boldsymbol{\mu}, \boldsymbol{\Sigma})$ are unique. The identifiability issue has been discussed in Liu et al. (2009). The uniqueness of $\boldsymbol{\mu}$ and $\boldsymbol{\Sigma}$ in $f(\boldsymbol{X})$ are imposed by modeling assumption: We assume that the transformation function $f$ preserves the first two marginal moments, i.e., $\mathbb{E}X_j = \mathbb{E}f_j(X_j)$ and $\text{Var}(X_j) = \text{Var}(f_j(X_j))$ for $j = 1, \ldots, d$. In this way, we can exploit the nonparanormal model in conducting the procedures that require more information besides the correlations.

## 2.3 Spearman's rho Correlation and Covariance Matrices

Given $n$ data points $\boldsymbol{x}_1, \ldots, \boldsymbol{x}_n \in \mathbb{R}^d$, where $\boldsymbol{x}_i = (x_{i1}, \ldots, x_{id})^T$, we denote by

$$\widehat{\mu}_j := \frac{1}{n} \sum_{i=1}^n x_{ij} \quad \text{and} \quad \widehat{\sigma}_j = \sqrt{\frac{1}{n} \sum_{i=1}^n (x_{ij} - \widehat{\mu}_j)^2},$$

the marginal sample means and standard deviations. Let $r_{ij}$ be the rank of $x_{ij}$ among $x_{1j}, \ldots, x_{nj}$ and $\bar{r}_j := \frac{1}{n} \sum_{i=1}^n r_{ij} = \frac{n+1}{2}$, we consider the following statistics:

$$\widehat{\rho}_{jk} = \frac{\sum_{i=1}^n (r_{ij} - \bar{r}_j)(r_{ik} - \bar{r}_k)}{\sqrt{\sum_{i=1}^n (r_{ij} - \bar{r}_j)^2 \cdot \sum_{i=1}^n (r_{ik} - \bar{r}_k)^2}},$$

and the correlation matrix estimators:

$$\widehat{\mathbf{R}}_{jk} = \begin{cases} 2 \sin\left(\frac{\pi}{6} \widehat{\rho}_{jk}\right) & j \neq k \\ 1 & j = k \end{cases}. \tag{2.1}$$

Equation (2.1) is inspired from Equation (6.4) in Kruskal (1958). We denote by $\widehat{\mathbf{R}} := [\widehat{\mathbf{R}}_{jk}]$ and $\widehat{\mathbf{S}} := [\widehat{\mathbf{S}}_{jk}] = [\widehat{\sigma}_j \widehat{\sigma}_k \widehat{\mathbf{R}}_{jk}]$ the Spearman's rho correlation and covariance matrices. Lemma 2.3, coming from Liu et al. (2012), claims that $\widehat{\mathbf{R}}$ can approach $\boldsymbol{\Sigma}^0$ in the parametric rate.

**Lemma 2.3** (Liu et al. (2012)). *When $\boldsymbol{x}_1, \ldots, \boldsymbol{x}_n \sim^{i.i.d} NPN_d(\boldsymbol{\Sigma}^0, f^0)$, for any $n \geq \frac{21}{\log d} + 2$, with probability at least $1 - 1/d^2$,*

$$\|\widehat{\mathbf{R}} - \boldsymbol{\Sigma}^0\|_{\max} \leq 8\pi \sqrt{\frac{\log d}{n}}. \tag{2.2}$$

# 3 Methods

In this section, we first provide the statistical models of Copula Component Analysis (COCA) and Copula PCA method. And then we introduce several algorithms to solve this problem.

## 3.1 Models

One of the intuition of PCA is coming from the Gaussian distribution. The principal components define the major axes of the contours of constant probability for the multivariate Gaussian (Anderson, 1958). However, such an interpretation does not exist when the distributions are away from the Gaussian. Balasubramanian and Schwartz (2002) constructed examples where PCA cannot preserve the structure of the data. Here we propose a toy example to show this phenomenon.

In Figure 1, we randomly generate 10,000 samples from three different types of nonparanormal distributions. We suppose that $\boldsymbol{X} \sim NPN_2(\boldsymbol{\Sigma}^0, f^0)$. Here we set $\boldsymbol{\Sigma}^0 = \begin{pmatrix} 1 & 0.5 \\ 0.5 & 1 \end{pmatrix}$ and transformation functions as follows: (A) $f_1^0(x) = x^3$ and $f_2^0(x) = x^{1/3}$; (B) $f_1^0(x) = \text{sign}(x)x^2$ and $f_2^0(x) = x^3$; (C) $f_1^0(x) = f_2^0(x) = \Phi^{-1}(x)$, where $\Phi$ is defined as the distribution function of the standard Gaussian distribution. Here, researchers might wish to conduct PCA separately on different clusters in (A) and (B). For (C), the data look very noisy and a nice major axis might be considered not existing.

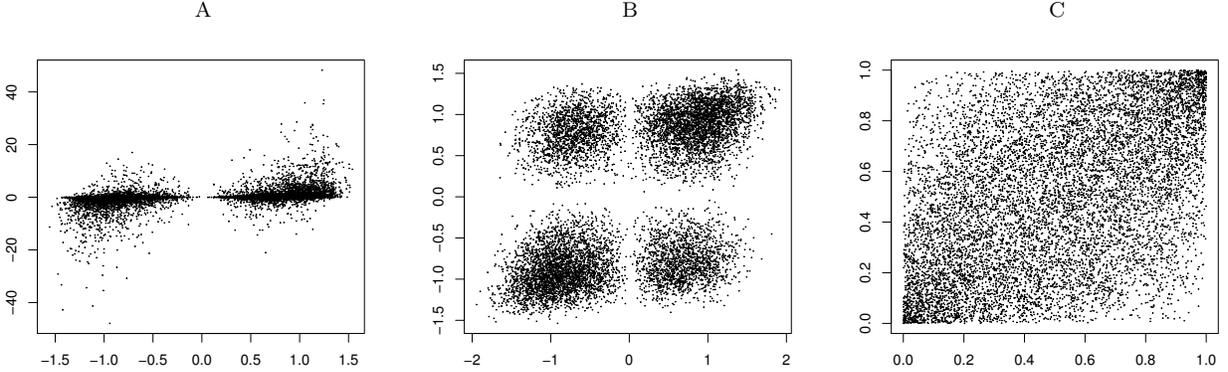

Figure 1: Scatter plots of three nonparanormals, $\boldsymbol{X} \sim NPN_2(\boldsymbol{\Sigma}^0, f^0)$. Here $\boldsymbol{\Sigma}^0_{12} = 0.5$ and the transformation functions have the form as follows: (A) $f^0_1(x) = x^3$ and $f^0_2(x) = x^{1/3}$; (B) $f^0_1(x) = \text{sign}(x)x^2$ and $f^0_2(x) = x^3$; (C) $f^0_1(x) = f^0_2(x) = \Phi^{-1}(x)$.

However, considering the monotone transformation $f^0$ as a type of data contamination, the geometric intuition of PCA comes back by estimating the principal components of the latent Gaussian distribution. In the next section, we will present the model of COCA and Copula PCA motivated from this observation.

### 3.1.1 COCA Model

We first show the model of Copula Component Analysis (COCA) method, where the idea of the scale-invariant PCA is exploited. We wish to estimate the leading eigenvector of the latent correlation matrix. In particular, let $\boldsymbol{\theta}_1$ be the leading eigenvectors of $\boldsymbol{\Sigma}^0$. For $0 \leq q \leq 1$, the $\ell_q$ ball $\mathbb{B}_q(R_q)$ is defined as:

$$\text{when} \quad q = 0, \quad \mathbb{B}_0(R_0) := \{\boldsymbol{v} \in \mathbb{R}^d : \text{card}(\text{supp}(\boldsymbol{v})) \leq R_0\};$$
$$\text{when} \quad 0 < q \leq 1, \quad \mathbb{B}_q(R_q) := \{\boldsymbol{v} \in \mathbb{R}^d : \|\boldsymbol{v}\|_q^q \leq R_q\}.$$

Accordingly, the COCA model $\mathcal{M}^0(q, R_q, \boldsymbol{\Sigma}^0, f^0)$ is considered:

$$\mathcal{M}^0(q, R_q, \boldsymbol{\Sigma}^0, f^0) = \{\boldsymbol{X} : \boldsymbol{X} \sim NPN_d(\boldsymbol{\Sigma}^0, f^0), \ \boldsymbol{\theta}_1 \in \mathbb{S}^{d-1} \cap \mathbb{B}_q(R_q)\}. \tag{3.1}$$

The $\ell_q$ ball induces a (weak) sparsity pattern when $0 \leq q \leq 1$ and has been analyzed in linear regression (Raskutti et al., 2011) and sparse PCA (Paul and Johnstone, 2012; Vu and Lei, 2012). Moreover, the data are assumed to come from a nonparanormal (or Gaussian copula) distribution, which contains and is a much larger distribution family than the Gaussian.

Inspired by the model $\mathcal{M}^0(q, R_q, \boldsymbol{\Sigma}^0, f^0)$, we consider the following estimator $\widetilde{\boldsymbol{\theta}}_1$, which is the global optimum to the following equation with the constraint that $\widetilde{\boldsymbol{\theta}}_1 \in \mathbb{B}_q(R_q)$ for some $0 \leq q \leq 1$:

$$\widetilde{\boldsymbol{\theta}}_1 = \arg\max_{\boldsymbol{v} \in \mathbb{R}^d} \ \boldsymbol{v}^T \widehat{\mathbf{R}} \boldsymbol{v}, \quad \text{subject to} \quad \boldsymbol{v} \in \mathbb{S}^{d-1} \cap \mathbb{B}_q(R_q). \tag{3.2}$$

Here $\widehat{\mathbf{R}}$ is the estimated Spearman's rho correlation matrix. The corresponding estimator $\widetilde{\boldsymbol{\theta}}_1$ can be considered as a nonlinear dimensional reduction procedure and has the potential to gain more flexibility compared with PCA.

### 3.1.2 Copula PCA Model

In contrast, we provide another method called Copula PCA, where we wish to estimate the leading eigenvector of the latent covariance matrix. In particular, let $\boldsymbol{u}_1$ be the leading eigenvector of $\boldsymbol{\Sigma}$. The following Copula PCA model $\mathcal{M}(q, R_q, \boldsymbol{\Sigma}, f)$ is considered:

$$\mathcal{M}(q, R_q, \boldsymbol{\Sigma}, f) = \big\{ \boldsymbol{X} : \boldsymbol{X} \sim MNPN_d(\boldsymbol{\mu}, \boldsymbol{\Sigma}, f), \ \ \boldsymbol{u}_1 \in \mathbb{S}^{d-1} \cap \mathbb{B}_q(R_q) \big\}. \quad (3.3)$$

An estimator corresponding to the above model is:

$$\widetilde{\boldsymbol{u}}_1 = \underset{\boldsymbol{v} \in \mathbb{R}^d}{\arg\max} \ \boldsymbol{v}^T \widehat{\mathbf{S}} \boldsymbol{v}, \quad \text{subject to} \quad \boldsymbol{v} \in \mathbb{S}^{d-1} \cap \mathbb{B}_q(R_q), \quad (3.4)$$

where $\widehat{\mathbf{S}}$ is the Spearman's rho covariance matrix.

### 3.1.3 Attainability of the Proposed Estimators

The direct computation of estimators $\widetilde{\boldsymbol{\theta}}_1$ and $\widetilde{\boldsymbol{u}}_1$ as defined in Equation (3.2) and (3.4) might be time consuming. However, in the following section we will show several algorithms which could approach these two global optimums and have good empirical performance. In particular, in Section 4 we will provide the theoretical performance in terms of guarantees of convergence and convergence rate of parameter estimation for the proposed algorithms. We will show that the global optimums proposed in Equations (3.2) and (3.4) can be well approached by using the truncated power algorithm. This algorithm has a (weaker) guarantee of convergence and under certain sufficient conditions the corresponding estimator can achieve the same convergence rate as the global optimum. Detailed theoretical analysis is provided in Section 4 as two new theorems (Theorems 4.7 and 4.9).

## 3.2 Algorithms

In this section we provide three sparse PCA algorithms, which the Spearman's rho correlation and covariance matrices $\widehat{R}$ and $\widehat{S}$ can be directly plugged in.

### 3.2.1 COCA and Copula PCA with PMD

Penalized Matrix Decomposition (PMD) is proposed by Witten et al. (2009). The main idea of PMD is a bi-convex optimization algorithm to the following problem:

$$\underset{\boldsymbol{v}, \boldsymbol{w}}{\arg\max} \ \boldsymbol{v}^T \widehat{\boldsymbol{\Gamma}} \boldsymbol{w}, \ \text{s.t.} \|\boldsymbol{v}\|_2^2 \leq 1, \|\boldsymbol{w}\|_2^2 \leq 1, \|\boldsymbol{v}\|_1 \leq \delta, \|\boldsymbol{w}\|_1 \leq \delta.$$

COCA with PMD and Copula PCA with PMD are listed in the following:

- 1. Input: A symmetric matrix $\widehat{\boldsymbol{\Gamma}}$. Initialize $\boldsymbol{w} \in \mathbb{S}^{d-1}$.

- 2. Iterate until convergence:
  (a) $\boldsymbol{v} \leftarrow \arg\max_{\boldsymbol{v} \in \mathbb{R}^d} \boldsymbol{v}^T \widehat{\boldsymbol{\Gamma}} \boldsymbol{w}$ subject to $\|\boldsymbol{v}\|_1 \leq \delta, \|\boldsymbol{v}\|_2^2 \leq 1$.
  (b) $\boldsymbol{w} \leftarrow \arg\max_{\boldsymbol{w} \in \mathbb{R}^d} \boldsymbol{v}^T \widehat{\boldsymbol{\Gamma}} \boldsymbol{w}$ subject to $\|\boldsymbol{w}\|_1 \leq \delta, \|\boldsymbol{w}\|_2^2 \leq 1$.
- 3. Output: $\boldsymbol{w}$.

Here $\widehat{\boldsymbol{\Gamma}}$ is either $\widehat{\boldsymbol{R}}$ or $\widehat{\boldsymbol{S}}$, corresponding to COCA with PMD and Copula PCA with PMD. $\delta$ is the tuning parameter. In practice, Witten et al. (2009) suggested using the first leading eigenvector of $\widehat{\boldsymbol{\Gamma}}$ to be the initial value. PMD can be considered as a solver to Equation (3.2) and Equation (3.4) with $q = 1$.

### 3.2.2 COCA and Copula PCA with SPCA

The SPCA algorithm is proposed by Zou et al. (2006). The main idea of SPCA is to exploit a regression approach to PCA and then utilize the lasso and elastic net (Zou and Hastie, 2005) to calculate a sparse estimator. COCA with SPCA and Copula PCA with SPCA are listed as follows:

- 1. Input: A symmetric matrix $\widehat{\boldsymbol{\Gamma}}$. Initialize $\boldsymbol{v} \in \mathbb{S}^{d-1}$.
- 2. Iterate until convergence:
  (a) $\boldsymbol{w} \leftarrow \arg\min_{\boldsymbol{w} \in \mathbb{R}^d}(\boldsymbol{v} - \boldsymbol{w})^T \widehat{\boldsymbol{\Gamma}}(\boldsymbol{v} - \boldsymbol{w}) + \delta_1 \|\boldsymbol{w}\|_2^2 + \delta_2 \|\boldsymbol{w}\|_1$;
  (b) $\boldsymbol{v} \leftarrow \widehat{\boldsymbol{\Gamma}} \boldsymbol{w} / \|\widehat{\boldsymbol{\Gamma}} \boldsymbol{w}\|_2$.
- 3. Output: $\boldsymbol{w}/\|\boldsymbol{w}\|_2$.

Here $\widehat{\boldsymbol{\Gamma}}$ is either $\widehat{\boldsymbol{R}}$ or $\widehat{\boldsymbol{S}}$, corresponding to COCA with SPCA and Copula PCA with SPCA. $\delta_1 \in \mathbb{R}$ and $\delta_2 \in \mathbb{R}$ are two tuning parameters. In practice, Zou et al. (2006) suggested using the first leading eigenvector of $\widehat{\boldsymbol{\Gamma}}$ to be the initial value. SPCA can also be considered as a solver to Equation (3.2) and Equation (3.4) with $q = 1$.

### 3.2.3 COCA and Copula PCA with TPower

Truncated power method (TPower) is proposed by Yuan and Zhang (2013). The main idea of TPower is to utilize the power method, but truncate the vector to a $\ell_0$ ball in each iteration. Actually, TPower can be generalized to a family of algorithms to solve Equation (3.2) when $0 \leq q \leq 1$, as presented in Algorithm 3.1. We name it $\ell_q$ Constraint Truncated Power Method (qTPM). In particular, when $q = 0$, the algorithm qTPM coincides with Yuan and Zhang (2013)'s method.

More specifically, we use the classical power method, but in each iteration $t$ we project the intermediate vector $\boldsymbol{x}_t$ to the intersection of the $d$-dimension sphere $\mathbb{S}^{d-1}$ and the $\ell_q$ ball with the radius $R_q^{1/q}$. The idea is to sort $\boldsymbol{x}_t$ from the highest to the lowest and find the highest $k$ absolute values and truncate all the others to zero, such that the resulting vector lies in $\mathbb{S}^{d-1} \cap \mathbb{B}_q(R_q)$ and is closest to the boundary of $\mathbb{B}_q(R_q)$.

For any vector $\boldsymbol{v} = (v_1, \ldots, v_d)^T$ and a index set $J \subset \{1, \ldots, d\}$, we define the truncation function TRC to be

$$\mathrm{TRC}(\boldsymbol{v}, J) := \big(v_1 \cdot I(1 \in J), \ \ldots, \ v_d \cdot I(d \in J)\big)^T, \tag{3.5}$$

where $I(\cdot)$ is the indicator function. Realizing that for any $p > q > 0$ and $\bm{v} \in \mathbb{R}^d$, $\|\bm{v}\|_p \leq \|\bm{v}\|_q \leq n^{1/q - 1/p}\|\bm{v}\|_p$, we have that the $\ell_q$ ball constraint is only active when $R_q \leq d^{1 - \frac{q}{2}}$. In practice, $R_q$ can be regarded as a tuning parameter. Lemma 3.1 states that, when $R_q > 1$, in each step of the iteration there exists a unique solution. In the following $a^{1/0} := a$ for any $a \in \mathbb{R}$.

---

**Algorithm 1** $\ell_q$ Constraint Truncated Power Method

---

**Input:** : symmetry matrix $\widehat{\bm{\Gamma}}$, initial vector $\widetilde{\bm{\theta}}_{q,0} \in \mathbb{R}^d$
**Output:** : $\widetilde{\bm{\theta}}_{q,\infty}$
  Let $t = 1$ and $R_q$ be the tuning parameter
  **repeat**
    compute $\bm{x}_t = \widehat{\bm{\Gamma}} \cdot \widetilde{\bm{\theta}}_{q,t-1} / \|\widehat{\bm{\Gamma}} \cdot \widetilde{\bm{\theta}}_{q,t-1}\|_2$
    **if** $\|\bm{x}_t\|_q \leq R_q^{1/q}$ **then**
      $\widetilde{\bm{\theta}}_{q,t} = \bm{x}_t$
    **else**
      Let $A_{tk}$ be the indices of $\bm{v}_t$ with the largest $k$ absolute values
      Compute $1 \leq k \leq d-1$ such that $\|\operatorname{TRC}(\bm{x}_t, A_{tk})/\|\operatorname{TRC}(\bm{x}_t, A_{tk})\|_2\|_q \leq R_q^{1/q}$ and $\|\operatorname{TRC}(\bm{x}_t, A_{t(k+1)})/\|\operatorname{TRC}(\bm{x}_t, A_{t(k+1)})\|_2\|_q > R_q^{1/q}$
      $\widetilde{\bm{\theta}}_{q,t} = \operatorname{TRC}(\bm{x}_t, A_{tk})/\|\operatorname{TRC}(\bm{x}_t, A_{tk})\|_2$
    **end if**
    $t \leftarrow t + 1$
  **until** Convergence

---

**Lemma 3.1.** Given $\bm{v} := (v_1, \ldots, v_d)^T$ with

$$v_1 \geq v_2 \geq \ldots \geq v_d \geq 0 \quad \text{and} \quad A_k = \{1, \ldots, k\}$$

the for any $0 < q \leq 1$ and $k \in \{1, \ldots, d-1\}$,

$$\frac{\|\bm{v}_{A_{k+1}}\|_q}{\|\bm{v}_{A_{k+1}}\|_2} \geq \frac{\|\bm{v}_{A_k}\|_q}{\|\bm{v}_{A_k}\|_2} \geq 1. \tag{3.6}$$

When $q = 0$, qTPM reduces to TPower algorithm proposed by Yuan and Zhang (2013). Therefore, we can combine COCA estimation consistency result in the next section with Theorem 1 in Yuan and Zhang (2013) to obtain a geometric convergence rate. Detailed theoretical analysis will be provided in Section 4. Because our main focus is on COCA instead of the sparse PCA algorithm, the general convergence rate for qTPM will be discussed in another paper. In practice, we will use the estimator obtained from SPCA (Zou et al., 2006) as the initial starting point, as suggested by Yuan and Zhang (2013).

### 3.2.4 Generalization to the First $m$ Sparse Eigenvectors

We use the iterative deflation method to learn the first $m$ instead of the first one leading eigenvectors, following the discussions of Mackey (2009), Journée et al. (2010), Yuan and Zhang (2013), and Zhang et al. (2012). In detail, a matrix $\widehat{\boldsymbol{\Gamma}} \in \mathbb{R}^{d \times s}$ deflates a vector $\boldsymbol{v} \in \mathbb{R}^d$ and results to a new matrix $\widehat{\boldsymbol{\Gamma}}'$:

$$\widehat{\boldsymbol{\Gamma}}' := (\mathbf{I} - \boldsymbol{v}\boldsymbol{v}^T)\widehat{\boldsymbol{\Gamma}}(\mathbf{I} - \boldsymbol{v}\boldsymbol{v}^T). \tag{3.7}$$

In this way, $\widehat{\boldsymbol{\Gamma}}'$ is orthogonal to $\boldsymbol{v}$.

### 3.2.5 Projection to the Positive Semi-definite Matrices Cone

To fit in the convex formulation in sparse PCA like semidefinite relaxation DSPCA (d'Aspremont et al., 2004), we project $\widehat{\mathbf{R}}$ into the cone of the positive semidefinite matrices and find solution $\widetilde{R}$ to the following convex optimization problem:

$$\widetilde{\mathbf{R}} = \underset{\mathbf{M} \succeq 0}{\arg\min} \|\widehat{\mathbf{R}} - \mathbf{M}\|_{\max}. \tag{3.8}$$

Here $\ell_{\max}$ norm is chosen such that the theoretical properties in Lemma 2.3 can be preserved. In particular, we have the following lemma:

**Lemma 3.2.** For all $t \geq 16\pi\sqrt{\frac{\log d}{n \log 2}}$, for any $n \geq \frac{37\pi}{t} + 2$, the minimizer $\widetilde{\mathbf{R}}$ to Equation (3.8) satisfies the following exponential inequality for all $1 \leq j, k \leq d$:

$$\mathbb{P}(|\widetilde{\mathbf{R}}_{jk} - \boldsymbol{\Sigma}^0_{jk}| \geq t) \leq 2\exp\left(-\frac{nt^2}{128\pi^2}\right). \tag{3.9}$$

In practice, the optimization problem in Equation (3.8) can be formulated as the dual of a graphical lasso problem with the smallest possible tuning parameter that still guarantees a feasible solution (Liu et al., 2012). And then we define $\widetilde{\mathbf{R}}$ and $\widetilde{\mathbf{S}} := [\widetilde{\mathbf{S}}_{jk}] = [\widehat{\sigma}_j \widehat{\sigma}_k \widetilde{R}_{jk}]$ to be the projected Spearman's rho correlation and covariance matrices. In practice we can always do such a projection and use $\widetilde{\mathbf{R}}$ and $\widetilde{\mathbf{S}}$ instead of $\widehat{\mathbf{R}}$ and $\widehat{\mathbf{S}}$.

## 4 Theoretical Properties

In this section we provide the theoretical properties of COCA and Copula PCA methods. In particular, we are interested in the high dimensional case when $d > n$ with both $d$ and $n$ increasing.

### 4.1 Rank-based Correlation and Covariance Matrices Estimation

In this section we state the main result on quantifying the convergence rate of $\widehat{\mathbf{R}}$ to $\boldsymbol{\Sigma}^0$ and $\widehat{\mathbf{S}}$ to $\boldsymbol{\Sigma}$. In particular, we establish the results on the $\ell_{\max}$ convergence rates of the Spearman's rho correlation and covariance matrices to $\boldsymbol{\Sigma}$ and $\boldsymbol{\Sigma}^0$.

For COCA, Lemma 2.3 is enough. For Copula PCA, however, we still need to quantify the convergence rate of $\widehat{\mathbf{S}}$ to $\boldsymbol{\Sigma}$. The key to prove the dominant eigenvector can be recovered in a fast rate is to show that the estimated covariance matrix $\widehat{\mathbf{S}}$ converges to $\boldsymbol{\Sigma}$ in the $\ell_{max}$ norm in a fast rate. To this end, we need extra conditions on the unknown transformation functions $\{f_j\}_{j=1}^d$. We define the *subgaussian transformation function class*. Let $(\sigma_1^2, \ldots, \sigma_d^2)^T := \text{diag}(\boldsymbol{\Sigma})$ and

$$\boldsymbol{\Sigma} = \begin{pmatrix} \sigma_1 & 0 & \ldots & 0 \\ 0 & \sigma_2 & \ldots & 0 \\ . & . & \ldots & . \\ 0 & 0 & \ldots & \sigma_d \end{pmatrix} \cdot \boldsymbol{\Sigma}^0 \cdot \begin{pmatrix} \sigma_1 & 0 & \ldots & 0 \\ 0 & \sigma_2 & \ldots & 0 \\ . & . & \ldots & . \\ 0 & 0 & \ldots & \sigma_d \end{pmatrix}.$$

**Definition 4.1.** Let $Z \in \mathbb{R}$ be a random variable following the standard Gaussian distribution. The subgaussian transformation function class $\text{TF}(K)$ is defined as the set of functions $\{g_0 : \mathbb{R} \to \mathbb{R}\}$ which satisfies that:

$$\mathbb{E}|g_0(Z)|^m \leq \frac{m!}{2} K^m, \quad \forall\, m \in \mathbb{Z}^+.$$

**Remark 4.1.** Here we note that for any function $g_0 : \mathbb{R} \to \mathbb{R}$, if there exists a constant $L < \infty$ such that

$$g_0(z) \leq L \quad \text{or} \quad g_0'(z) \leq L \quad \text{or} \quad g_0''(z) \leq L, \quad \forall\, z \in \mathbb{R}, \tag{4.1}$$

then $g_0 \in \text{TF}(K)$ for some constant K. To show that, we have the central absolute moments of the standard Gaussian distribution satisfying, $\forall\, m \in \mathbb{Z}^+$:

$$\mathbb{E}|Z|^m \leq (m-1)!! < m!! \quad \text{and} \quad \mathbb{E}|Z^2|^m = (2m-1)!! < m! \cdot 2^m. \tag{4.2}$$

Because $g_0$ satisfies the condition in Equation (4.1), using Taylor expansion, we have for any $z \in \mathbb{R}$,

$$g_0(z) \leq |g_0(0)| + L \text{ or } |g_0(z)| \leq |g_0(0)| + L|z|, \text{ or } |g_0(z)| \leq |g_0(0)| + |g_0'(0)z| + Lz^2. \tag{4.3}$$

Combining Equations (4.2) and (4.3), we have $\mathbb{E}|g_0(Z)|^m \leq \frac{m!}{2} K^m$ for some constant $K$. This proves the assertion.

Then we have the following result, which states that $\boldsymbol{\Sigma}$ can also be recovered in the parametric rate. The key of the proof is to show that the marginal sample means and standard deviations of the nonparanormal can converge to the population means and standard deviations in an exponential rate.

**Lemma 4.2.** Let $\boldsymbol{x}_1, \ldots, \boldsymbol{x}_n$ be $n$ independent realizations of a random vector $\boldsymbol{X}$, where $\boldsymbol{X} \sim MNPN_d(\boldsymbol{\mu}, \boldsymbol{\Sigma}, f)$. If $g := \{g_j = f_j^{-1}\}_{j=1}^d$ satisfies for all $j = 1, \ldots, K$, $g_j^2 \in TF(K)$ where $K < \infty$ is some constant, we have for any $1 \leq j, k \leq d$, for any $n \geq \frac{37\pi}{t} + 2$,

$$\mathbb{P}(|\widehat{\mathbf{S}}_{jk} - \boldsymbol{\Sigma}_{jk}| > t) \leq 2 \exp(-c_1 n t^2), \tag{4.4}$$
$$\mathbb{P}(|\widehat{\mu}_j - \mu_j| > t) \leq 2 \exp(-c_2 n t^2), \tag{4.5}$$

where $c_1$ and $c_2$ are two constants only depending on the choice of $K$.

**Remark 4.3.** Lemma 4.2 claims that, under certain constraint on the transformation functions, the latent covariance matrix $\boldsymbol{\Sigma}$ can be recovered using the Spearman's rho covariance matrix. However, in this case, the marginal distributions of the nonparanormal are required to be sub-gaussian and cannot be arbitrarily continuous. This makes Copula PCA a less favored method compared with COCA.

## 4.2 COCA and Copula PCA

In this section we provide the main result on the upper bound of the estimation error of COCA estimators and Copula PCA estimators. We say that the model $\mathcal{M}^0(q, R_q, \boldsymbol{\Sigma}^0, f^0)$ holds if the data are drawn from an element in the model $\mathcal{M}^0(q, R_q, \boldsymbol{\Sigma}^0, f^0)$; We say that the model $\mathcal{M}(q, R_q, \boldsymbol{\Sigma}, f)$ holds if the data are drawn from an element in the model $\mathcal{M}(q, R_q, \boldsymbol{\Sigma}, f)$.

The next theorem provides an upper bound on the angle between the global optimum $\widetilde{\boldsymbol{\theta}}_1$ to Equation (3.2) and the true parameter $\boldsymbol{\theta}_1$.

**Theorem 4.4.** Let $\widetilde{\boldsymbol{\theta}}_1$ be the global optimum in Equation (3.2) and the model $\mathcal{M}^0(q, R_q, \boldsymbol{\Sigma}^0, f^0)$ holds. For any two vectors $\boldsymbol{v}_1 \in \mathbb{S}^{d-1}$ and $\boldsymbol{v}_2 \in \mathbb{S}^{d-1}$, let $|\sin \angle(\boldsymbol{v}_1, \boldsymbol{v}_2)| := \sqrt{1 - (\boldsymbol{v}_1^T \boldsymbol{v}_2)^2}$. Then we have, for any $n \geq \frac{21}{\log d} + 2$, with probability at least $1 - 1/d^2$,

$$\sin^2 \angle(\widetilde{\boldsymbol{\theta}}_1, \boldsymbol{\theta}_1) \leq \gamma_q R_q^2 \left( \frac{64\pi^2}{(\lambda_1 - \lambda_2)^2} \cdot \frac{\log d}{n} \right)^{\frac{2-q}{2}},$$

where $\gamma_q = 2 \cdot I(q = 1) + 4 \cdot I(q = 0) + (1 + \sqrt{3})^2 \cdot I(0 < q < 1)$ and $\lambda_j = \Lambda_j(\boldsymbol{\Sigma}^0)$ for $j = 1, 2$.

*Proof.* The key idea of the proof is to utilize the $\ell_{\max}$ norm convergence result of $\widehat{\mathbf{R}}$ to $\boldsymbol{\Sigma}^0$ as shown in Lemma 2.3, then apply the proof of Theorem 2.2 in Vu and Lei (2012). For self-containedness, a proof is provided in Section B.4 □

It can be observed that the convergence rate of $\widetilde{\boldsymbol{\theta}}_1$ to $\boldsymbol{\theta}_1$ will be faster when $\boldsymbol{\theta}_1$ lies in a more sparse ball. It makes sense because the effect of "the curse of dimensionality" will be decreasing when the parameters are more and more sparse. Generally, when $R_q$ and $\lambda_1, \lambda_2$ do not scale with $(n, d)$, the rate is $O_P\left((\frac{\log d}{n})^{1-q/2}\right)$, which is the parametric rate Ma (2013), Vu and Lei (2012), and Paul and Johnstone (2012) obtained.

Given Theorem 4.4, we can immediately obtain the following corollary, which quantifies the expected angle between $\widetilde{\boldsymbol{\theta}}_1$ and $\boldsymbol{\theta}_1$.

**Corollary 4.1.** In the conditions of Theorem 4.4, we have

$$\mathbb{E} \sin^2 \angle(\widetilde{\boldsymbol{\theta}}_1, \boldsymbol{\theta}_1) \leq \gamma_q R_q^2 \left( \frac{64\pi^2}{(\lambda_1 - \lambda_2)^2} \cdot \frac{\log d}{n} \right)^{\frac{2-q}{2}} + \frac{1}{d^2}.$$

*Proof.* Define $\epsilon = \sin \angle(\widetilde{\boldsymbol{\theta}}_1, \boldsymbol{\theta}_1)$. Because $\sin^2(\cdot) \in [0, 1]$, using Theorem 4.4, we have

$$\mathbb{E}\epsilon^2 = \mathbb{E}\left[\epsilon^2 I\left(\epsilon^2 \leq \gamma_q R_q^2 \left(\frac{64\pi^2}{(\lambda_1-\lambda_2)^2} \cdot \frac{\log d}{n}\right)^{\frac{2-q}{2}}\right)\right] + \mathbb{E}\left[\epsilon^2 I\left(\epsilon^2 > \gamma_q R_q^2 \left(\frac{64\pi^2}{(\lambda_1-\lambda_2)^2} \cdot \frac{\log d}{n}\right)^{\frac{2-q}{2}}\right)\right]$$

$$\leq \gamma_q R_q^2 \left(\frac{64\pi^2}{c_1(\lambda_1-\lambda_2)^2} \cdot \frac{\log d}{n}\right)^{\frac{2-q}{2}} + \frac{1}{d^2}.$$

This completes the proof. □

In the next corollary, we provide a sparsity recovery consistency result for $\widetilde{\boldsymbol{\theta}}_1$. It can be observed that the true sparsity pattern can be recovered in a fast rate given a constraint on the minimum absolute value of the signal part of $\boldsymbol{\theta}_1$.

**Corollary 4.2.** Let $\widetilde{\boldsymbol{\theta}}_1$ be the global solution to Equation (3.2) and the model $\mathcal{M}^0(0, R_0, \boldsymbol{\Sigma}^0, f^0)$ holds. Let $\Theta^0 := \mathrm{supp}(\boldsymbol{\theta}_1)$ and $\widehat{\Theta}^0 := \mathrm{supp}(\widetilde{\boldsymbol{\theta}}_1)$. If we further have $\min_{j \in \Theta^0} |\theta_{1j}| \geq \frac{16\sqrt{2}R_0\pi}{\lambda_1 - \lambda_2}\sqrt{\frac{\log d}{n}}$, then for any $n \geq \frac{21}{\log d} + 2$, $\mathbb{P}(\widehat{\Theta}^0 = \Theta^0) \geq 1 - d^{-2}$.

*Proof.* The key of the proof is to construct a contradiction given Theorem 4.4 and the condition on the minimum absolute value of nonzero entries of $\boldsymbol{\theta}_1$. Detailed proof can be found in Section B.5. □

Similarly, we can give an upper bound for the estimation rate of the Copula PCA estimator $\widetilde{\boldsymbol{u}}_1$ to the true leading eigenvalue $\boldsymbol{u}_1$ of the latent covariance matrix $\boldsymbol{\Sigma}$. The next theorem provides the detail result.

**Theorem 4.5.** Let $\widetilde{\boldsymbol{u}}_1$ be the global solution to Equation (3.4) and the model $\mathcal{M}(q, R_q, \boldsymbol{\Sigma}, f)$ holds. If $g := \{g_j = f_j^{-1}\}_{j=1}^d$ satisfies $g_j^2 \in TF(K)$ for all $1 \leq j \leq d$, then we have, for any $n \geq \frac{21}{\log d} + 2$, with probability at least $1 - 1/d^2$,

$$\sin^2 \angle(\widetilde{\boldsymbol{u}}_1, \boldsymbol{u}_1) \leq \gamma_q R_q^2 \left( \frac{4}{c_1(\omega_1 - \omega_2)^2} \cdot \frac{\log d}{n} \right)^{\frac{2-q}{2}},$$

where $\gamma_q = 2 \cdot I(q=1) + 4 \cdot I(q=0) + (1+\sqrt{3})^2 \cdot I(0 < q < 1)$, $\omega_j = \Lambda_j(\boldsymbol{\Sigma})$ for $j=1,2$ and $c_1$ is a constant defined in Equation (4.4), only depending on $K$.

*Proof.* Under the conditions that $g := \{g_j = f_j^{-1}\}_{j=1}^d$ satisfies $g_j^2 \in TF(K)$ for all $1 \leq j \leq d$, we can utilize Lemma 4.2 and have that

$$\mathbb{P}(|\widehat{\mathbf{S}}_{jk} - \boldsymbol{\Sigma}_{jk}| > t) \leq 2\exp(-c_1 n t^2), \quad \forall\, j, k \in \{1, \ldots, d\}.$$

Using this key observation, all the proofs in Theorem 4.4 can still proceed until Equation (B.14). In particular, let $\epsilon_u := \sin\angle(\boldsymbol{u}_1, \widetilde{\boldsymbol{u}}_1)$, we have

$$\mathbb{P}(\epsilon_u^2 \geq t) \leq \mathbb{P}\left( \frac{\gamma_q R_q^2}{(\omega_1 - \omega_2)^{2-q}} \|\mathrm{vec}(\widehat{\mathbf{S}} - \boldsymbol{\Sigma})\|_\infty^{2-q} \geq t \right) = \mathbb{P}\left( \|\widehat{\mathbf{S}} - \boldsymbol{\Sigma}\|_{\max} \geq \left( \frac{t(\omega_1 - \omega_2)^{2-q}}{\gamma_q R_q^2} \right)^{1/(2-q)} \right)$$

$$\leq d^2 \exp\left( -c_1 n \left( \frac{t(\lambda_1 - \lambda_2)^{2-q}}{\gamma_q R_q^2} \right)^{2/(2-q)} \right).$$

Choosing $t = \gamma_q R_q^2 \left( \frac{4}{c_1(\omega_1 - \omega_2)^2} \frac{\log d}{n} \right)^{\frac{2-q}{2}}$, we have the result. □

Given Theorem 4.5, we can immediately obtain the following corollary, which bounds the expected angle between $\widetilde{\boldsymbol{u}}_1$ and $\boldsymbol{u}_1$.

**Corollary 4.3.** In the conditions of Theorem 4.5, we have

$$\mathbb{E}\sin^2 \angle(\widetilde{\boldsymbol{u}}_1, \boldsymbol{u}_1) \leq \gamma_q R_q^2 \left( \frac{4}{c_1(\omega_1 - \omega_2)^2} \cdot \frac{\log d}{n} \right)^{\frac{2-q}{2}} + \frac{1}{d^2}.$$

*Proof.* Using the same techniques in proving Corollary 4.1. □

Similarly, we can prove that under mild conditions $\widetilde{\boldsymbol{u}}_1$ can recover the support set of $\boldsymbol{u}_1$.

**Corollary 4.4.** Let $\widetilde{\boldsymbol{u}}_1$ be the global solution to Equation (3.4) and the model $\mathcal{M}(0, R_0, \boldsymbol{\Sigma}, f)$ holds. Let $\Theta := \mathrm{supp}(\boldsymbol{u}_1)$ and $\widehat{\Theta} := \mathrm{supp}(\widetilde{\boldsymbol{u}}_1)$. If $g := \{g_j = f_j^{-1}\}_{j=1}^d$ satisfies $g_j^2 \in TF(K)$ for all $1 \leq j \leq d$ and we further have $\min_{j \in \Theta} |u_{1j}| \geq \frac{4\sqrt{2}R_0}{\sqrt{c_1}(\omega_1 - \omega_2)} \sqrt{\frac{\log d}{n}}$, then for any $n \geq \frac{21}{\log d} + 2$, $\mathbb{P}(\widehat{\Theta} = \Theta) \geq 1 - \frac{1}{d^2}$.

*Proof.* Using the same techniques in proving Corollary 4.2. □

**Remark 4.6.** Assuming that the transformation function $g$ satisfies that $g_j^2 \in TF(K)$ for $j = 1, \ldots, d$ restricts the distribution families of the nonparanormal. We note that this constraint is close to claiming that the marginal distributions of the random vector $\boldsymbol{X}$ have sub-gaussian tails. However, Copula PCA is still an interesting procedure in estimating the leading eigenvectors in the sense that it provides a sparse PCA approach on a model strictly larger than the Gaussian, while consistently and robustly estimating the true latent leading eigenvector in a fast rate.

Let $\breve{\boldsymbol{\theta}}_1$ denote the estimator derived using the Truncated Power method, as shown in Algorithm 3.1 by setting $q = 0$ and the input matrix $\boldsymbol{\Gamma}$ to be $\widehat{\mathbf{R}}$. In the next theorem we show that, under mild conditions, $\breve{\boldsymbol{\theta}}_1$ can approach $\boldsymbol{\theta}_1$ in a fast near-optimal rate.

**Theorem 4.7.** Let the tuning parameter in TPower be denoted by $k := \mathrm{card}(\mathrm{supp}(\breve{\boldsymbol{\theta}}_1))$ such that $k \geq 4R_0$ and the initial starting point be denoted by $\boldsymbol{v}_0$ with $\mathrm{card}(\mathrm{supp}(\boldsymbol{v}_0)) \leq k$ and $\|\boldsymbol{v}_0\|_2 = 1$. Let

$$\nu_1 := \frac{n\lambda_2 + 8\pi(R_0 + 2k)\sqrt{\log d}}{n\lambda_1 - 8\pi(R_0 + 2k)\sqrt{\log d}}$$

and $\nu_2 := \frac{8\sqrt{2}\pi(R_0 + 2k)\sqrt{\log d}}{\sqrt{n(\lambda_1 - \lambda_2)^2 - 32\pi(\lambda_1 - \lambda_2)(R_0 + 2k)\sqrt{n \log d} + 320\pi^2(R_0 + 2k)^2 \log d}}.$

If the model $\mathcal{M}^0(0, R_0, \boldsymbol{\Sigma}^0, f^0)$ holds and the following three assumptions hold:

(A1) $\lambda_1$ and $\lambda_2$ scale with $(n, d)$ such that $\lambda_1 - \lambda_2 \geq 16\pi(R_0 + 2k)\sqrt{\frac{\log d}{n}}$;

(A2) $(1 + 3\sqrt{R_0/k})(1 - 0.45(1 - \nu_1^2)) < 1$;

(A3) Letting $\zeta_1 := |\boldsymbol{\theta}_1^T \boldsymbol{v}_0| - \nu_2$ be a fixed constant in $[0, 1]$, we have $0 < (1 - \nu_1^2)\zeta_1(1 - \zeta_1^2)/2 - 2\nu_2 - \sqrt{R_0/k} < 1$,

We have, with probability larger than $1 - d^{-2}$,

$$|\sin \angle(\breve{\boldsymbol{\theta}}_1, \boldsymbol{\theta}_1)| \leq \frac{C}{\lambda_1 - \lambda_2}(R_0 + 2k) \cdot \sqrt{\frac{\log d}{n}},$$

for some generic constant $C$ not scaled with $(n, d)$.

*Proof.* The key of the proof is to show that

$$\max_{\boldsymbol{v}\in\mathbb{S}^{d-1}\cap\mathbb{B}_0(R_0+2k)} |\boldsymbol{v}^T(\widehat{\mathbf{R}}-\boldsymbol{\Sigma}^0)\boldsymbol{v}| \leq 8\pi(R_0+2k)\sqrt{\frac{\log d}{n}}$$

with large probability and for any $\boldsymbol{v}_1, \boldsymbol{v}_2 \in \mathbb{S}^{d-1}$,

$$\sqrt{1-|\boldsymbol{v}_1^T\boldsymbol{v}_2|} \leq |\sin\angle(\boldsymbol{v}_1,\boldsymbol{v}_2)| \leq 2\sqrt{1-|\boldsymbol{v}_1^T\boldsymbol{v}_2|}.$$

Detailed proof can be found in Section B.6. □

**Remark 4.8.** Here Assumption (A1) is to control the difference between the top two leading eigenvalues $\lambda_1$ and $\lambda_2$, such that $\boldsymbol{\theta}_1$ can be differentiated from $\boldsymbol{\theta}_2$. Assumption (A3) is to control the closedness of the initial value $\boldsymbol{v}_0$ to $\boldsymbol{\theta}_1$. This assumption makes sense because Truncated Power method is a nonconvex formulation in estimating $\boldsymbol{\theta}_1$. The theory verifies that, when assumptions hold, the obtained estimator $\widecheck{\boldsymbol{\theta}}_1$ can obtain the same convergence rate as the global optimum $\widetilde{\boldsymbol{\theta}}_1$.

Let $\widecheck{\boldsymbol{u}}_1$ denote the estimator derived using the Truncated Power method(TPower), as shown in Algorithm 3.1 by setting $q = 0$ and the input matrix $\boldsymbol{\Gamma}$ to be $\widehat{\mathbf{S}}$. In the next theorem we show that, under mild conditions, $\widecheck{\boldsymbol{u}}_1$ can approach $\boldsymbol{u}_1$ in a fast near-optimal rate.

**Theorem 4.9.** Let the tuning parameter in TPower be denoted by $k := \text{card}(\text{supp}(\widecheck{\boldsymbol{u}}_1))$ such that $k \geq 4R_0$ and the initial starting point be denoted by $\boldsymbol{v}_0$ with $\text{card}(\text{supp}(\boldsymbol{v}_0)) \leq k$ and $\|\boldsymbol{v}_0\|_2 = 1$. Let

$$\nu_3 := \frac{n\lambda_2\sqrt{c_1} + 2(R_0+2k)\sqrt{\log d}}{n\lambda_1\sqrt{c_1} - 2(R_0+2k)\sqrt{\log d}}$$

$$\text{and} \quad \nu_4 := \frac{2\sqrt{2}(R_0+2k)\sqrt{\log d}}{\sqrt{nc_1(\lambda_1-\lambda_2)^2 - 8(\lambda_1-\lambda_2)(R_0+2k)\sqrt{nc_1\log d} + 20(R_0+2k)^2\log d}}.$$

If the model $\mathcal{M}(0, R_0, \boldsymbol{\Sigma}, f)$ holds and the following three assumptions hold:

(B1) $\omega_1$ and $\omega_2$ scale with $(n,d)$ such that $\omega_1 - \omega_2 \geq 4(R_0+2k)\sqrt{\frac{\log d}{nc_1}}$;

(B2) $(1+3\sqrt{R_0/k})(1-0.45(1-\nu_3^2)) < 1$;

(B3) Letting $\zeta_2 := |\boldsymbol{u}_1^T\boldsymbol{v}_0| - \nu_4$ be a fixed constant in $[0,1]$, we have $0 < (1-\nu_3^2)\zeta_2(1-\zeta_2^2)/2 - 2\nu_4 - \sqrt{R_0/k} < 1$,

We have, with probability larger than $1 - d^{-2}$,

$$|\sin\angle(\widecheck{\boldsymbol{u}}_1, \boldsymbol{u}_1)| \leq \frac{C}{\omega_1 - \omega_2}(R_0+2k)\cdot\sqrt{\frac{\log d}{n}},$$

for some generic constant $C$ not scaled with $(n,d)$.

## 4.3 Discussion on the Attainability of the Optimum

In Section 3.1.3 we show that the optimums to Equations (3.2) and (3.4) are hard to compute. To approximate the global optimum $\widetilde{\boldsymbol{\theta}}_1$ and $\widetilde{\boldsymbol{u}}_1$, we advocate using the Truncated Power method (Yuan and Zhang, 2013) and provide the theoretical analysis for the corresponding algorithm, as shown in Theorems 4.7 and 4.9. To guarantee convergence of the proposed algorithm, we need to make sure that the initial vector $\boldsymbol{v}_0$ is not too far away from the true vector $\boldsymbol{\theta}_1$ or $\boldsymbol{u}_1$. In this section we discuss two approaches in finding such an vector $\boldsymbol{v}_0$ in light of the arguments in Yuan and Zhang (2013):

(i) As suggested by Yuan and Zhang (2013) (Paragraph 2, Page 905), to find a proper initial vector $\boldsymbol{v}_0$, we can take a relatively large pilot tuning parameter $\bar{k}$ so that the requirement on $\boldsymbol{\theta}_1^T \boldsymbol{v}_0 \gtrsim \sqrt{R_0/\bar{k}}$ is easier to be satisfied. Using $\bar{k}$ we get a pilot estimator $\bar{\boldsymbol{v}}$ and then plug it into the qTPM algorithm with a smaller tuning parameter $k$. Yuan and Zhang (2013) provided some theoretical justification for this procedure. They also provided thorough numerical experiments to show that this approach is practically effective in applications.

(ii) An alternative way to choose the initial vector $\boldsymbol{v}_0$ is to exploit the estimator obtained from other sparse PCA algorithms to initialize qTPM. For example, we can plug the Spearman's rho correlation and covariance matrices $\widehat{\mathbf{R}}$ and $\widehat{\mathbf{S}}$ into the Sparse PCA algorithm with the semidefinite programming formulation (d'Aspremont et al., 2004) (We call it the SDP algorithm). From the theory of Yuan and Zhang (2013), we know that if the SDP procedure provides a consistent estimator of $\boldsymbol{\theta}_1$, we could use the SDP estimator to initialize the qTPM algorithm and achieve the desired rate.

## 4.4 Discussion on the Optimal Rate of Convergence of COCA

Many results have been established in understanding the sparse PCA problem. For example, under the Gaussian assumptions, Amini and Wainwright (2009) discuss the problem of support recovery of leading eigenvectors, Berthet and Rigollet (2013) discuss the problem of sparse principal component detection and Vu and Lei (2012) propose methods that obtain a $\sqrt{R_0 \log d / n}$ rate of convergence for parameter estimation when $\boldsymbol{u}_1$ are sparse with support set size $R_0$ and show that this rate is minimax optimal confined in the Gaussian family.

COCA is significantly different from the procedures in the above mentioned papers in the sense that: (i) With regard to methodology, we suggest using the Spearman's rho correlation matrix $\widehat{\mathbf{R}}$ to estimate $\boldsymbol{\Sigma}^0$, instead of using the sample correlation matrix $\mathbf{S}^0$. Empirical results in the next section show that rank-based methods is more robust to modeling and data contaminations than the methods based on the Pearson sample correlation matrix. (ii) With regard to theory, in terms of modeling flexibility, COCA gains main compared with the results in Ma (2013), Vu and Lei (2012), and Paul and Johnstone (2012): The nonparanormal family contains many heavy-tailed distributions with arbitrary margins, which cannot be handled by the Gaussian-based procedures. COCA is the optimal method when $R_0$ is fixed. When not, it is unclear whether COCA is the optimal method confined in the nonparanormal family.

Addressing the optimal rate of convergence of COCA is challenging due to the reason that the data can be very heavy-tailed and the transformed rank-based correlation matrix has a much more complex structure than the Pearson's covariance/correlation matrix.

However, here we lay out a venue in attempt to prove a sharper rate of convergence of COCA. More specifically, we prove that COCA can attain the parametric $\sqrt{s\log d/n}$ rate of convergence if a condition called "third-order sign subgaussian condition" holds for the nonparanormally distributed random vector $\boldsymbol{X}$.

**Definition 4.2** (third-order sign subgaussian condition). Let $\boldsymbol{X}_1$ be a random vector and $\boldsymbol{X}_2, \boldsymbol{X}_3$ be two independent copies of $\boldsymbol{X}_1$. For any random vector $\boldsymbol{v} \in \mathbb{S}^{d-1}$, we let $\mathbf{O} \in \mathcal{R}^{d \times d}$ be the population-wise Spearman's rho matrix with

$$\mathbf{O} := 3 \cdot \mathbb{E}\left\{\mathrm{sign}(\boldsymbol{X}_1 - \boldsymbol{X}_2)(\mathrm{sign}(\boldsymbol{X}_1 - \boldsymbol{X}_3))^T\right\},$$

and let

$$Y_{\boldsymbol{v}} := \boldsymbol{v}^T \mathrm{sign}(\boldsymbol{X}_1 - \boldsymbol{X}_2)(\mathrm{sign}(\boldsymbol{X}_1 - \boldsymbol{X}_3))^T \boldsymbol{v}.$$

Then $\boldsymbol{X}_1$ is said to satisfy the third-order sign subgaussian condition if and only if there exists an absolute constant $c$ such that for any $\boldsymbol{v} \in \mathbb{S}^{d-1}$,

$$\mathbb{E}\exp\{t(Y_{\boldsymbol{v}} - \mathbb{E}Y_{\boldsymbol{v}})\} \leq \exp(c(\|\boldsymbol{\Sigma}_0\|_2 + \|\mathbf{O}\|_2)^2 t^2), \quad \text{for } |t| < t_0, \qquad (4.6)$$

where $t_0$ is a positive number such that $t_0(\|\boldsymbol{\Sigma}_0\|_2 + \|\mathbf{O}\|_2)^2$ is lower bounded by a fixed constant.

We then have the following theorem, which states that we can recover $\boldsymbol{\theta}_1$ in the parametric rate of convergence when $\boldsymbol{X}$ satisfies the third-order sign subgaussian condition.

**Theorem 4.10.** When the model $\mathcal{M}^0(0, R_0, \boldsymbol{\Sigma}^0, f^0)$ holds and the nonprarnomally distributed random ector $\boldsymbol{X}$ satisfies Equation (4.6), we have

$$|\sin\angle(\widetilde{\boldsymbol{\theta}}_1, \boldsymbol{\theta}_1)| = O_P\left(\frac{\lambda_1 + \|\mathbf{O}\|_2}{\lambda_1 - \lambda_2}\sqrt{\frac{R_0 \log d}{n}}\right).$$

Theorem 4.10 can be shown to be correct in three steps and we sketch the proof as follows.

(i) By using the argument in Liu et al. (2012) (Page 2319), we have

$$\widehat{\rho}_{jk} = \frac{n-2}{n-1}U_{jk} + \frac{3}{n+1}\widehat{\tau}_{jk},$$

where $\widehat{\tau}_{jk} \in [-1, 1]$ is the Kendall's tau correlation coefficient and

$$U_{jk} = \frac{3}{n(n-1)(n-2)}\sum_{i \neq s \neq t}\mathrm{sign}(x_{ij} - x_{sj})(x_{ik} - x_{tk}).$$

(ii) We only focus on $U_{jk}$ and then following the proof of Lemma 5.4 in Han and Liu (2013) until Equation (5.21), where we substitute Equation (5.22) by (4.6), we can prove that

$$\|\widehat{\mathbf{O}} - \mathbf{O}\|_2 = O_P\left((\lambda_1 + \|\mathbf{O}\|_2)\sqrt{\frac{R_0 \log d}{n}}\right),$$

where $\widehat{\mathbf{O}}$ is the empirical realization of $\mathbf{O}$ with $\widehat{\mathbf{O}}_{jk} = \widehat{\rho}_{jk}$ for $j, k \in \{1, \ldots, d\}$.

(iii) Combining with the proof of Lemma C.2 in Wegkamp and Zhao (2013), we can show that the $\sin(\cdot)$ transformation in $\widehat{\mathbf{R}}$ does not hurt the rate and hence we have

$$\|\widehat{\mathbf{R}} - \boldsymbol{\Sigma}_0\|_2 = O_P\left((\lambda_1 + \|\mathbf{O}\|_2)\sqrt{\frac{R_0 \log d}{n}}\right).$$

This completes the proof.

## 5 Experiments

In this section we investigate the empirical performance of the COCA method. Three sparse PCA algorithms are considered: Penalized Matrix Decomposition (PMD) proposed by Witten et al. (2009), SPCA proposed by Zou et al. (2006) and Truncated Power method (TPower) proposed by Yuan and Zhang (2013). The following three methods are considered:

- Pearson: the sparse PCA algorithm using the Pearson sample correlation matrix;
- Spearman: the sparse PCA algorithm using the Spearman's rho correlation matrix;
- Oracle: the sparse PCA algorithm using the Pearson sample correlation matrix of the latent Gaussian data (perfect without data contamination).

### 5.1 Numerical Simulations

In the simulation study we study the empirical performance for support recovery and parameter estimation for different estimators where samples are drawn from an element of the model $\mathcal{M}^0(0, R_0, \Sigma^0, f^0)$.

In detail, we sample $n$ data points $x_1, \ldots, x_n$ from the nonparanormal distribution $\boldsymbol{X} \sim NPN_d(\Sigma^0, f^0)$. Here we set $d = 100$. We follow the same generating scheme as in Shen and Huang (2008) and Yuan and Zhang (2013). A covariance matrix $\Sigma$ is firstly synthesized through the eigenvalue decomposition, where the first two eigenvalues are given and the corresponding eigenvectors are pre-specified to be sparse. In detail, we suppose that the first two dominant eigenvectors of $\Sigma$, $u_1$ and $u_2$, are sparse in the sense that only the first $s = 10$ entries of $u_1$ and the second $s = 10$ entries of $u_2$ nonzero, i.e.,

$$u_{1j} = \begin{cases} \frac{1}{\sqrt{10}} & 1 \leq j \leq 10 \\ 0 & \text{otherwise} \end{cases} \text{ and } u_{2j} = \begin{cases} \frac{1}{\sqrt{10}} & 11 \leq j \leq 20 \\ 0 & \text{otherwise} \end{cases},$$

and $\omega_1 = 5, \omega_2 = 2, \omega_3 = \ldots = \omega_d = 1$. The remaining eigenvectors are chosen arbitrarily. The correlation matrix $\Sigma^0$ is accordingly generated from $\Sigma$, with $\lambda_1 = 4$, $\lambda_2 = 2.5$, $\lambda_3, \ldots, \lambda_d \leq 1$ and the two dominant eigenvectors sparse:

$$\theta_{1j} = \begin{cases} \frac{-1}{\sqrt{10}} & 1 \leq j \leq 10 \\ 0 & \text{otherwise} \end{cases} \text{ and } \theta_{2j} = \begin{cases} \frac{-1}{\sqrt{10}} & 11 \leq j \leq 20 \\ 0 & \text{otherwise} \end{cases}.$$

To sample data from the nonparanormal distribution, we also need the transformation functions: $f^0 = \{f_j^0\}_{j=1}^d$. Here two types of transformation functions are considered:

- Linear transformation (or no transformation):

$$f_{\text{linear}}^0 = \{h_0, h_0, \ldots, h_0\}, \quad \text{where} \quad h_0(x) := x.$$

- Nonlinear transformation: there exist five univariate monotone functions $h_1, h_2, \ldots, h_5 : \mathbb{R} \to \mathbb{R}$ and

$$f_{\text{nonlinear}}^0 = \{h_1, h_2, h_3, h_4, h_5, h_1, h_2, h_3, h_4, h_5, \ldots\},$$

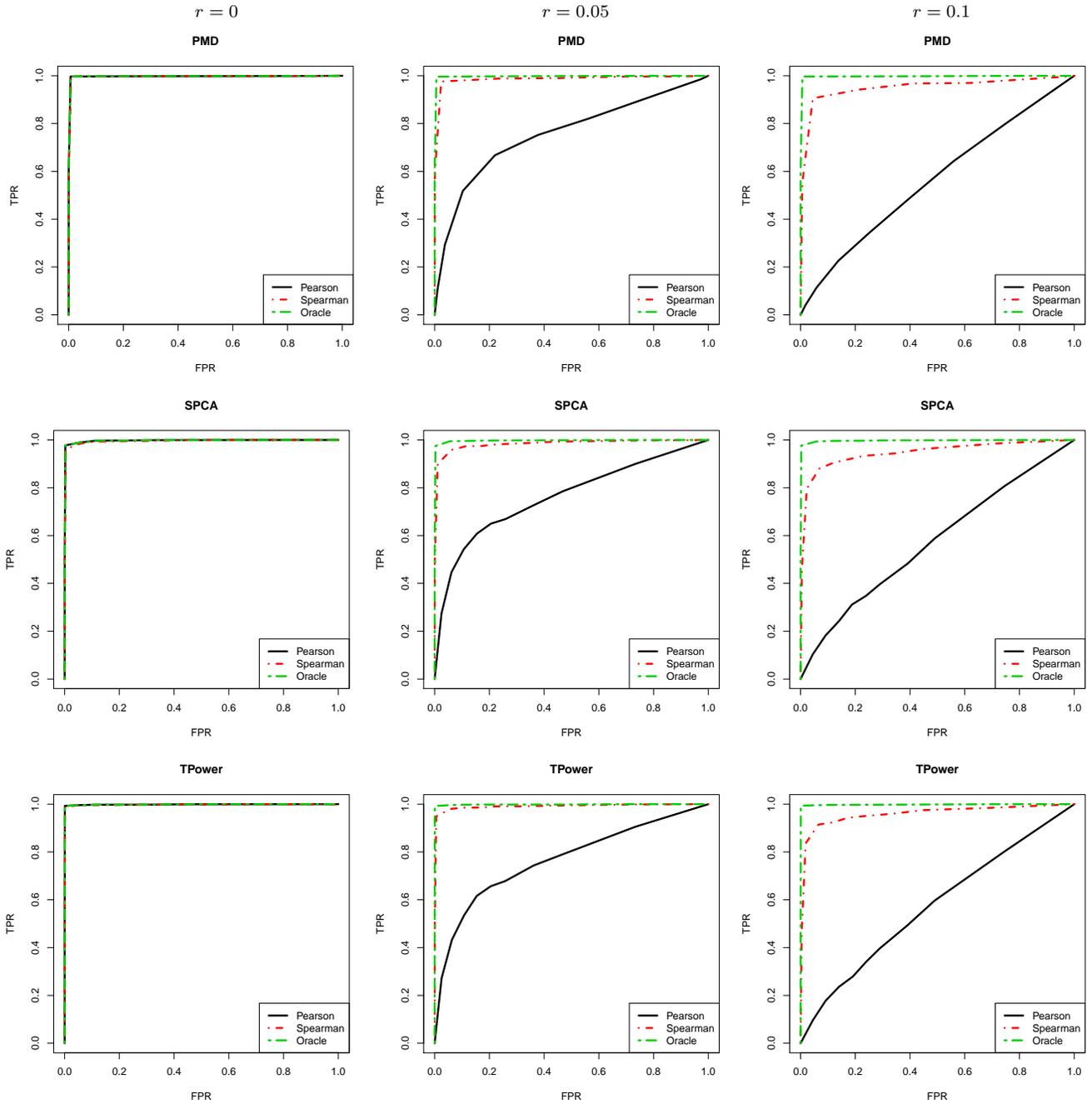

Figure 2: ROC curves for PMD, SPCA and Truncated Power method (top, middle, bottom) with linear (no) transformation and data contamination at different levels ($r = 0, 0.05, 0.1$). Here $n = 100$ and $d = 100$.

where

$$h_1^{-1}(x) := x, \ h_2^{-1}(x) := \frac{\text{sign}(x)|x|^{1/2}}{\sqrt{\int |t|\phi(t)dt}}, \ h_3^{-1}(x) := \frac{\Phi(x) - \int \Phi(t)\phi(t)dt}{\sqrt{\int \left(\Phi(y) - \int \Phi(t)\phi(t)dt\right)^2 \phi(y)dy}},$$

$$h_4^{-1}(x) := \frac{x^3}{\sqrt{\int t^6 \phi(t)dt}}, \ \text{and} \ h_5^{-1}(x) := \frac{\exp(x) - \int \exp(t)\phi(t)dt}{\sqrt{\int \left(\exp(y) - \int \exp(t)\phi(t)dt\right)^2 \phi(y)dy}}.$$

Here $\phi$ and $\Phi$ are defined to be the probability density and cumulative distribution functions of the standard Gaussian. We then generate $n = 100, 200$ or $500$ data points from:

- **[Scheme 1]** $\boldsymbol{X} \sim NPN_d(\boldsymbol{\Sigma}^0, f^0_{\text{linear}})$ where $f^0_{\text{linear}} = \{h_0, h_0, \ldots, h_0\}$ and $\boldsymbol{\Sigma}_0$ is defined as above.

- **[Scheme 2]** $\boldsymbol{X} \sim NPN_d(\boldsymbol{\Sigma}^0, f^0_{\text{nonlinear}})$ where $f^0_{\text{nonlinear}} = \{h_1, h_2, h_3, h_4, h_5, \ldots\}$ and $\boldsymbol{\Sigma}_0$ is defined as above.

To evaluate the robustness of different methods, we adopt a similar data contamination procedure as in Liu et al. (2012). Let $r \in [0, 1)$ represent the proportion of samples being contaminated. For each dimension, we randomly select $\lfloor nr \rfloor$ entries and replace them with either 5 or -5 with equal probability. The final data matrix we obtained is $\boldsymbol{X} \in \mathbb{R}^{n \times d}$. PMD, SPCA and TPower are then employed on $\boldsymbol{X}$ to computer the estimated leading eigenvector $\widetilde{\boldsymbol{\theta}}_1$.

To evaluate the empirical variable selection property of different methods, we define

$$\mathcal{S} := \{1 \leq j \leq d : \theta_{1j} \neq 0\} \ \text{and} \ \widehat{\mathcal{S}}_\delta := \{1 \leq j \leq d : \widetilde{\boldsymbol{\theta}}_{1j} \neq 0\},$$

to be the support sets of the true leading eigenvector $\theta_1$ and the estimated leading eigenvector $\widetilde{\boldsymbol{\theta}}_1$ using the tuning parameter $\delta$. In this way, the False Positive Number (FPN) and False Negative Number (FNN) of $\delta$ are defined as:

$$\text{FPN}(\delta) := \text{the number of features in } \widehat{\mathcal{S}}_\delta \text{ not in } \mathcal{S},$$
$$\text{FNN}(\delta) := \text{the number of features in } \mathcal{S} \text{ not in } \widehat{\mathcal{S}}_\delta.$$

Then we can further define the False Positive Rate(FPR) and False Negative Rate (FNR) corresponding to the tuning parameter $\delta$ to be

$$\text{FPR}(\delta) := \text{FPN}(\delta)/(d-s) \ \text{and} \ \text{FNR}(\delta) := \text{FNN}(\delta)/s.$$

Under the Scheme 1 and Scheme 2 with different levels of contamination ($r = 0, 0.05$ or $0.1$), we repeatedly generate the data matrix $\boldsymbol{X}$ for 1,000 times and compute the averaged False Positive Rates and False Negative Rates using a path of tuning parameters $\delta$. The feature selection performances of different methods are then evaluated by plotting $(\text{FPR}(\delta), 1 - \text{FNR}(\delta))$. The corresponding ROC curves are presented in Figure 2 and Figure 3.

In Figure 2, Scheme 1 is explored and it can be observed that under the most ideal case where there is no contamination ($r = 0$) and $\boldsymbol{X}$ is exactly Gaussian, Pearson, Spearman and Oracle can all recover the sparsity pattern perfectly.

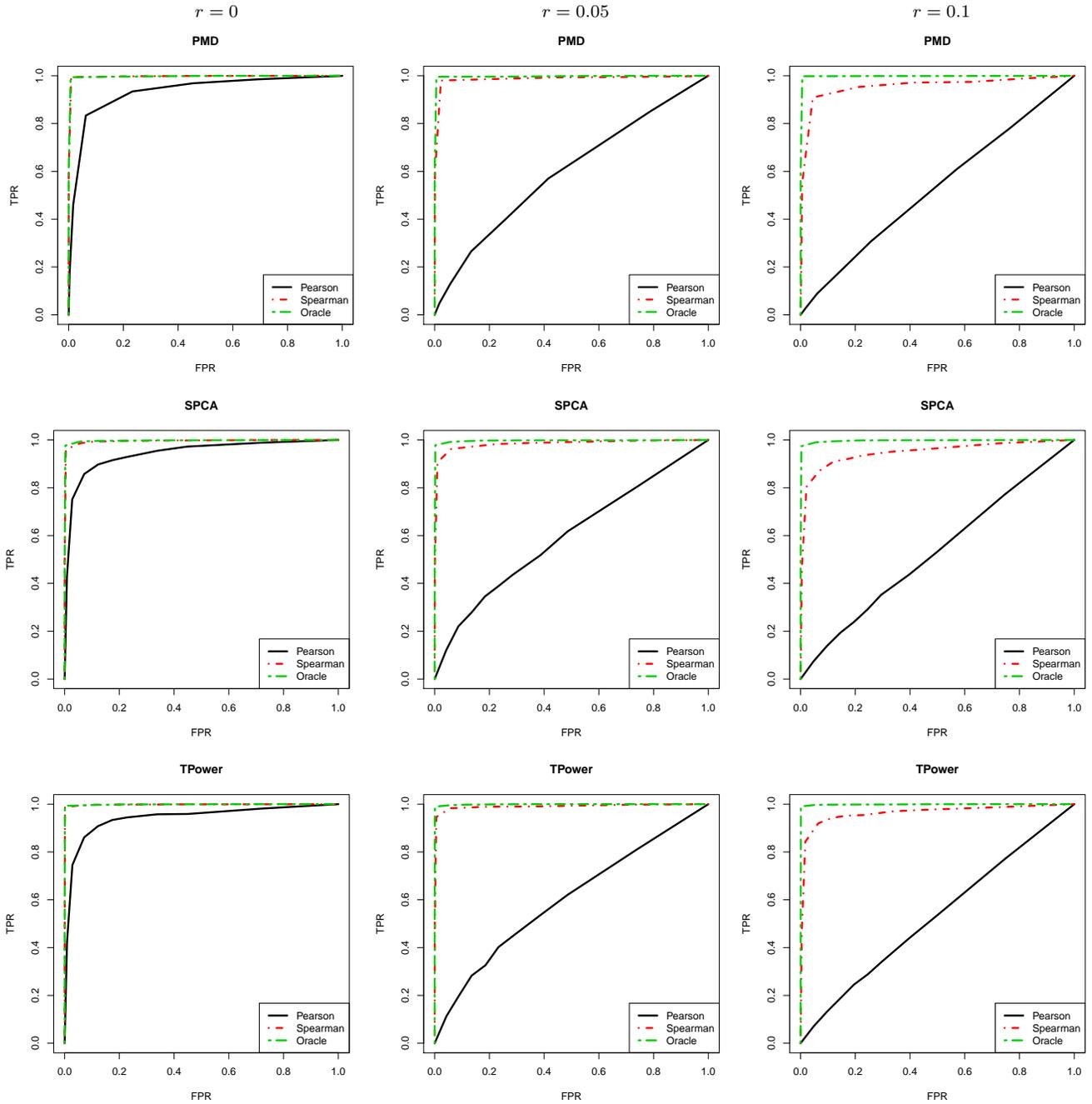

Figure 3: ROC curves for PMD, SPCA and Truncated Power method (top, middle, bottom) with nonlinear transformation and data contamination at different levels ($r = 0, 0.05, 0.1$). Here $n = 100$ and $d = 100$.

Table 1: Quantitative comparison on the dataset under the generating scheme 1 with linear transformation. The means of the $\sin \angle(\boldsymbol{\theta}_1, \widetilde{\boldsymbol{\theta}}_1)$ with their standard deviations in parentheses are presented. Here $n$ is changing from 100 to 500 and $d = 100$.

| method | n | r | Pearson | Spearman | Oracle |
|---|---|---|---|---|---|
| PMD | 100 | 0.00 | 0.2739(0.0337) | 0.2806(0.0372) | 0.2739(0.0337) |
|  |  | 0.05 | 0.6909(0.1307) | 0.3720(0.0941) | 0.2942(0.0620) |
|  |  | 0.10 | 0.9007(0.0852) | 0.4414(0.0952) | 0.2742(0.0293) |
|  | 200 | 0.00 | 0.2576(0.0000) | 0.2577(0.0012) | 0.2576(0.0000) |
|  |  | 0.05 | 0.4630(0.0969) | 0.2610(0.0110) | 0.2576(0.0000) |
|  |  | 0.10 | 0.7768(0.1089) | 0.2871(0.0409) | 0.2576(0.0000) |
|  | 500 | 0.00 | 0.2576(0.0000) | 0.2576(0.0000) | 0.2576(0.0000) |
|  |  | 0.05 | 0.2730(0.0234) | 0.2576(0.0000) | 0.2576(0.0000) |
|  |  | 0.10 | 0.4651(0.1033) | 0.2576(0.0000) | 0.2576(0.0000) |
| SPCA | 100 | 0.00 | 0.3686(0.0879) | 0.3952(0.0885) | 0.3686(0.0879) |
|  |  | 0.05 | 0.6765(0.0910) | 0.4412(0.0867) | 0.3605(0.0814) |
|  |  | 0.10 | 0.8660(0.0800) | 0.5173(0.0857) | 0.3614(0.0977) |
|  | 200 | 0.00 | 0.1869(0.0489) | 0.2060(0.0534) | 0.1869(0.0489) |
|  |  | 0.05 | 0.4335(0.0892) | 0.2451(0.0753) | 0.1836(0.0523) |
|  |  | 0.10 | 0.7016(0.0998) | 0.3236(0.0863) | 0.1874(0.0558) |
|  | 500 | 0.00 | 0.0762(0.0178) | 0.0833(0.0190) | 0.0762(0.0178) |
|  |  | 0.05 | 0.2319(0.0676) | 0.1045(0.0274) | 0.0807(0.0225) |
|  |  | 0.10 | 0.3854(0.0925) | 0.1362(0.0305) | 0.0799(0.0199) |
| TPower | 100 | 0.00 | 0.1126(0.0726) | 0.1312(0.0913) | 0.1126(0.0726) |
|  |  | 0.05 | 0.6513(0.1175) | 0.2423(0.1452) | 0.1132(0.0668) |
|  |  | 0.10 | 0.8726(0.0776) | 0.3900(0.1551) | 0.1096(0.0637) |
|  | 200 | 0.00 | 0.0709(0.0151) | 0.0761(0.0169) | 0.0709(0.0151) |
|  |  | 0.05 | 0.3730(0.1433) | 0.0933(0.0281) | 0.0683(0.0176) |
|  |  | 0.10 | 0.7310(0.0912) | 0.1306(0.0714) | 0.0667(0.0172) |
|  | 500 | 0.00 | 0.0424(0.0114) | 0.0459(0.0112) | 0.0424(0.0114) |
|  |  | 0.05 | 0.1210(0.0423) | 0.0581(0.0120) | 0.0420(0.0096) |
|  |  | 0.10 | 0.3858(0.1349) | 0.0694(0.0167) | 0.0422(0.0116) |

Table 2: Quantitative comparison on the dataset under the generating scheme 2 with nonlinear transformation. The means of the $\sin \angle(\boldsymbol{\theta}_1, \widetilde{\boldsymbol{\theta}}_1)$ with their standard deviations in parentheses are presented. Here $n$ is changing from 100 to 500 and $d = 100$.

| method | n | r | Normal | Spearman | Oracle |
|---|---|---|---|---|---|
| PMD | 100 | 0.00 | 0.5076(0.1504) | 0.2878(0.0451) | 0.2778(0.0361) |
|  |  | 0.05 | 0.8729(0.1025) | 0.3497(0.0820) | 0.2814(0.0421) |
|  |  | 0.10 | 0.9514(0.0584) | 0.4338(0.0952) | 0.2775(0.0371) |
|  | 200 | 0.00 | 0.3272(0.0743) | 0.2576(0.0000) | 0.2576(0.0000) |
|  |  | 0.05 | 0.6867(0.1359) | 0.2610(0.0139) | 0.2576(0.0000) |
|  |  | 0.10 | 0.8910(0.0919) | 0.2807(0.0370) | 0.2576(0.0000) |
|  | 500 | 0.00 | 0.2582(0.0036) | 0.2576(0.0000) | 0.2576(0.0000) |
|  |  | 0.05 | 0.4439(0.1096) | 0.2576(0.0000) | 0.2576(0.0000) |
|  |  | 0.10 | 0.7055(0.1421) | 0.2576(0.0000) | 0.2576(0.0000) |
| SPCA | 100 | 0.00 | 0.5210(0.0961) | 0.4005(0.0946) | 0.3768(0.1000) |
|  |  | 0.05 | 0.8453(0.0973) | 0.4470(0.0851) | 0.3673(0.0819) |
|  |  | 0.10 | 0.9245(0.0742) | 0.5141(0.0949) | 0.3556(0.0977) |
|  | 200 | 0.00 | 0.3583(0.0889) | 0.1949(0.0532) | 0.1788(0.0489) |
|  |  | 0.05 | 0.6448(0.1050) | 0.2729(0.0782) | 0.1847(0.0534) |
|  |  | 0.10 | 0.8502(0.1097) | 0.3212(0.0852) | 0.1927(0.0599) |
|  | 500 | 0.00 | 0.1744(0.0483) | 0.0843(0.0252) | 0.0780(0.0218) |
|  |  | 0.05 | 0.3699(0.0979) | 0.1053(0.0257) | 0.0788(0.0206) |
|  |  | 0.10 | 0.5546(0.1229) | 0.1318(0.0318) | 0.0779(0.0174) |
| TPower | 100 | 0.00 | 0.4516(0.1216) | 0.1346(0.0832) | 0.1202(0.0746) |
|  |  | 0.05 | 0.8315(0.1094) | 0.2372(0.1517) | 0.1053(0.0513) |
|  |  | 0.10 | 0.9323(0.0730) | 0.3608(0.1583) | 0.1088(0.0629) |
|  | 200 | 0.00 | 0.1942(0.1056) | 0.0740(0.0191) | 0.0702(0.0190) |
|  |  | 0.05 | 0.6193(0.1263) | 0.0900(0.0313) | 0.0661(0.0172) |
|  |  | 0.10 | 0.8608(0.0926) | 0.1266(0.0596) | 0.0663(0.0185) |
|  | 500 | 0.00 | 0.1025(0.0310) | 0.0465(0.0101) | 0.0437(0.0086) |
|  |  | 0.05 | 0.3296(0.1293) | 0.0586(0.0154) | 0.0422(0.0101) |
|  |  | 0.10 | 0.6296(0.1157) | 0.0708(0.0171) | 0.0403(0.0099) |

However, when the data are contaminated where outliers exist, the performances of Pearson utilizing PMD, SPCA and TPower significantly decrease, while the rank-based method Spearman is still very close to Oracle.

In Figure 3, Scheme 2 is explored and $\mathbf{X}$ follows a nonparanormal distribution and is non-Gaussian. It can be observed that, in Scheme 2, even without data contamination ($r = 0$), Pearson cannot recover the support set of $\boldsymbol{\theta}_1$, while textsfSpearman can still recover the sparsity pattern almost perfectly. When the data are contaminated where outliers exist, the performance of the rank-based method Spearman utilizing PMD, SPCA and TPower is still very close to Oracle.

To explore the empirical performances of difference methods using different algorithms more, we define an oracle tuning parameter $\delta^*$ to be the $\delta$ with the lowest $\text{FPR}(\delta) + \text{FNR}(\delta)$: $\delta^* := \arg\min_\delta (\text{FPR}(\delta) + \text{FNR}(\delta))$. In this way, an estimator $\widetilde{\boldsymbol{\theta}}_1$ using the oracle tuning parameter $\delta^*$ can be calculated and we computer the angle between $\boldsymbol{\theta}_1$ and $\widetilde{\boldsymbol{\theta}}_1$: $\sin\angle(\boldsymbol{\theta}_1, \widetilde{\boldsymbol{\theta}}_1)$ to quantify the estimation consistency.

In Table 1 and Table 2, the averaged $\sin\angle(\boldsymbol{\theta}_1, \widetilde{\boldsymbol{\theta}}_1)$ values for Scheme 1 and Scheme 2, $n = 100, 200, 500$, contamination levels $r = 0, 0.05, 0.1$ and utilizing three algorithms (PMD, SPCA and TPower) are presented. There are mainly three observations drawn from the results:

- In the perfectly Gaussian data (Scheme 1 with $r = 0$), Pearson performs slightly better than Spearman. However, the difference is not significantly. When $r \neq 0$, Spearman outperforms Pearson significantly.

- In Scheme 2 where the data are non-Gaussian, even when $r = 0$ and $n$ is large, Pearson's estimation error is significantly away from zero. Spearman can still achieve good performance here and perform much more robustly when $r \neq 0$ compared with Pearson.

- In both Scheme 1 and Scheme 2, when $r = 0$, Spearman is close to Oracle and is tending to zero when $n$ is large. When $r \neq 0$, the performance of Spearman drops, but significantly less than Pearson.

With regard to the comparison among the three algorithms (PMD, SPCA, TPower), we have two more comments restricted to what we observe:

- PMD's estimator $\widetilde{\boldsymbol{\theta}}_1$ seems not converging to $\boldsymbol{\theta}_1$ in our simulation studies. This might be due to the fact that PMD is more sensitive to the choice of initial values.

- TPower performs generally better than SPCA. We also find that the computing time of TPower is less than SPCA.

## 5.2 Large-scale Genomic Data Analysis

In this section we investigate the performance of Spearman compared with Pearson using one of the largest microarray datasets (McCall et al., 2010). In summary, we collect in all 13,182 publicly available microarray samples from Affymetrixs HGU133a platform. The raw data contain 20,248 probes and 13,182 samples belonging to 2,711 tissue types (e.g., lung cancers, prostate cancer, brain tumor etc.). There are at most 1599 samples and at least 1 sample belonging to each tissue type. We merge the probes corresponding to the same gene. There are remaining 12,713 genes and 13,182 samples. The main purpose of this experiment is to compare the performance of Spearman with Pearson. We use the Truncated Power method proposed by Yuan and Zhang (2013) in this section.

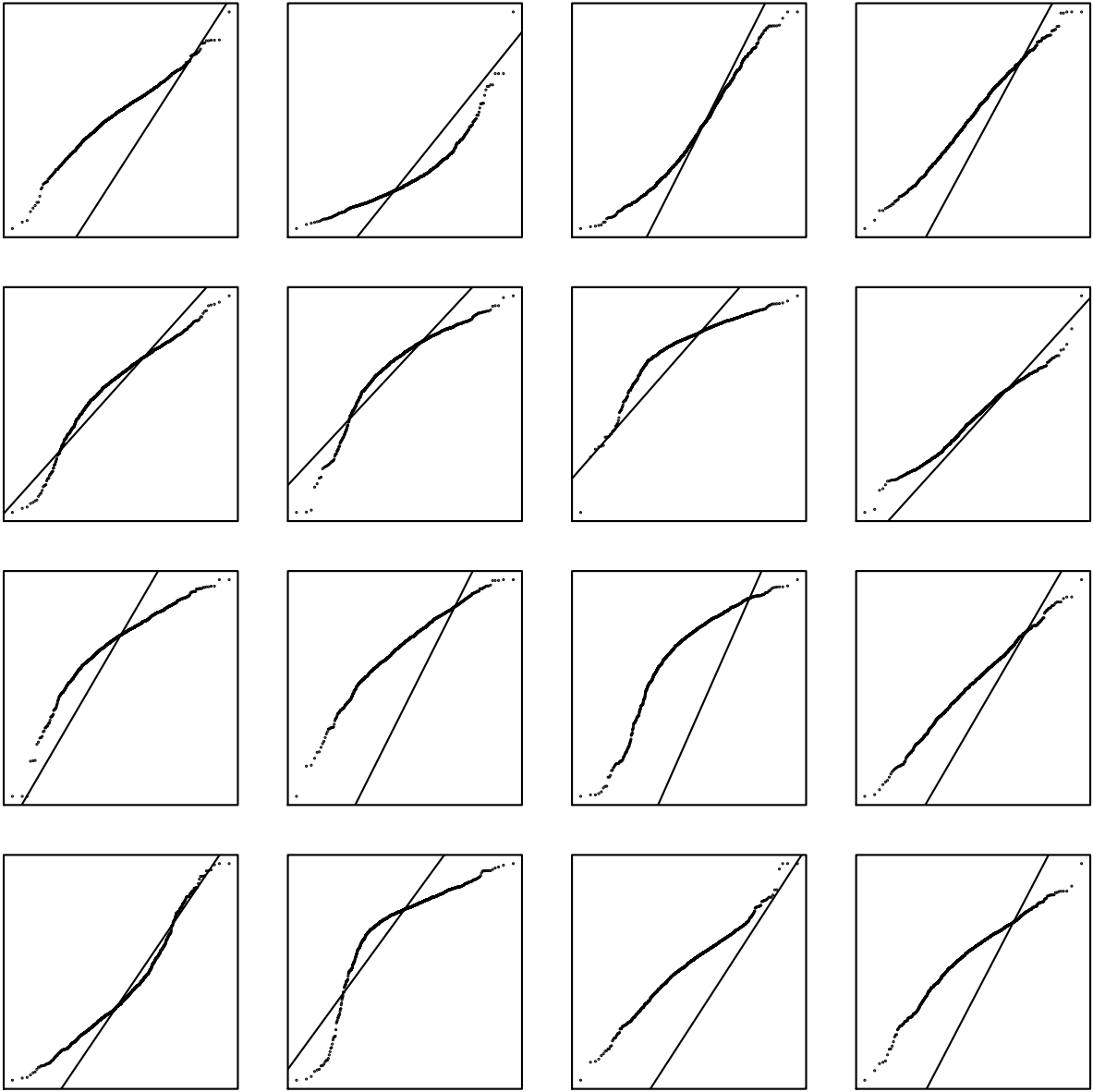

Figure 4: Sixteen randomly picked genes' Quantile-to-Quantile plots. The x-axis represents the theoretical quantiles and the y-axis represents the sample quantiles.

We first show that the data are non-Gaussian. To this end, we randomly pick 16 genes and all samples from a certain tissue type, then the corresponding Quantile-to-Quantile plots (QQ plots) compared with the Gaussian are presented in Figure 4 to illustrate their normality. It can be observed that all the sixteen marginal distributions are severely away from the Gaussian.

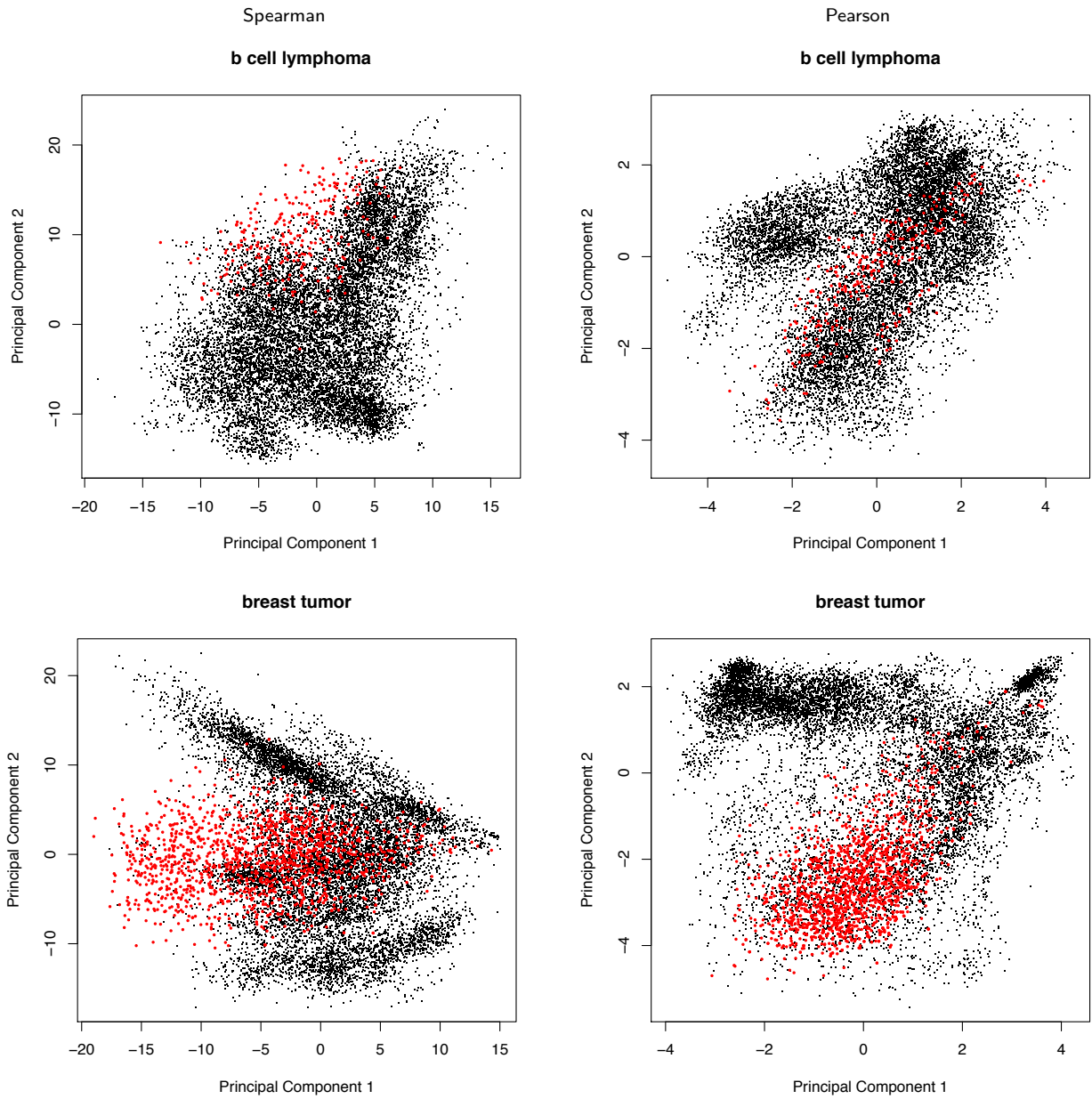

Figure 5: The scatter plots of the first two principal components of the dataset. Spearman and Pearson are compared (left to right) and b cell lymphoma and breast tumor are explored (top to bottom). Each black point represents a sample and each red point represents a sample belonging to the corresponding tissue type.

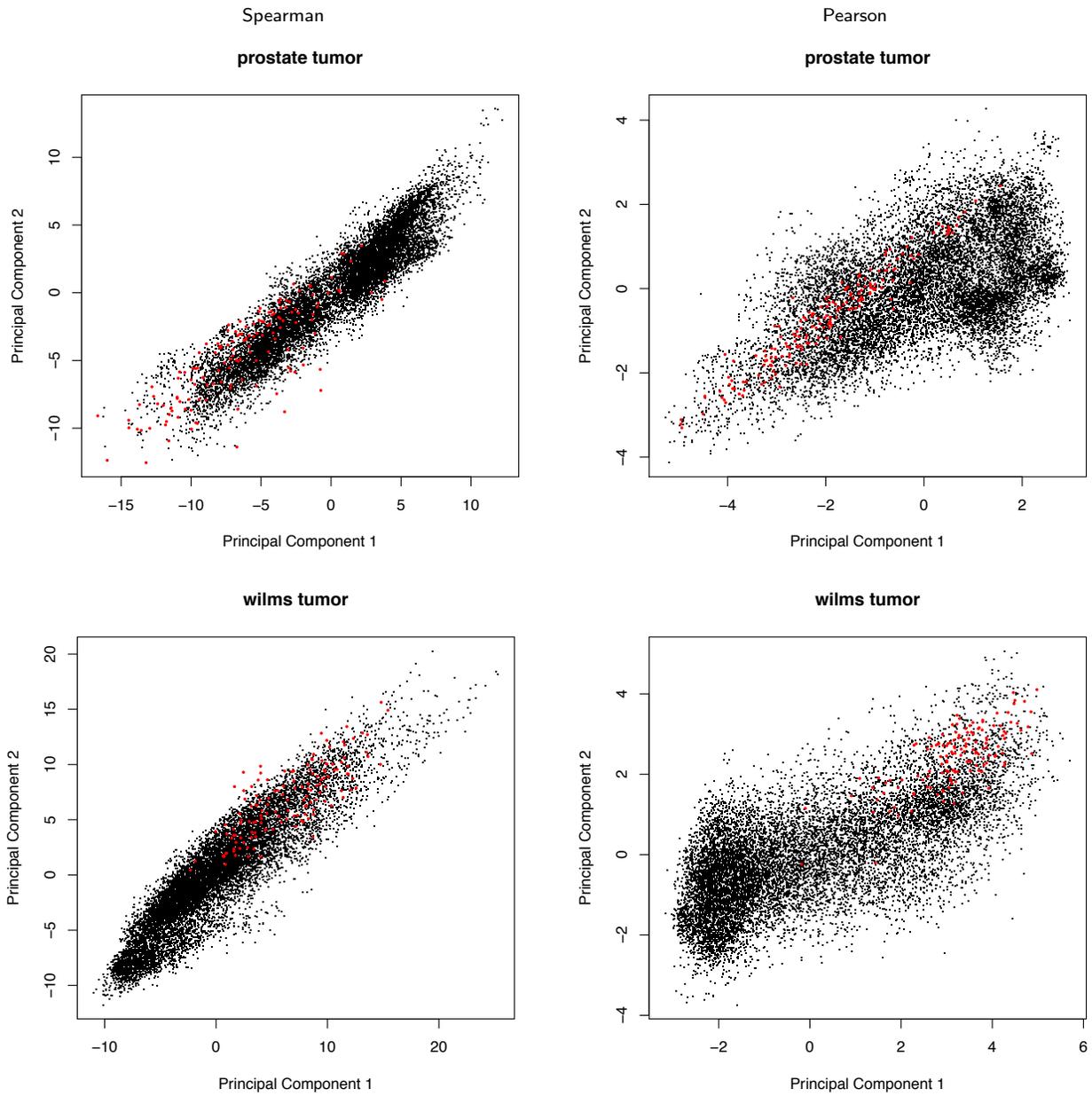

Figure 6: The scatter plots of the first two principal components of the dataset. Spearman and Pearson are compared (left to right) and prostate tumor and Wilms tumor are explored (top to bottom). Each black point represents a sample and each red point represents a sample belonging to the corresponding tissue type.

We adopt the same idea of data-preprocessing as in Liu et al. (2012). In particular, we firstly remove the batch effect by applying the surrogate variable analysis proposed by Leek and Storey (2007). We then extract the top 2,000 genes with the highest marginal standard deviations. There are, accordingly, 2,000 genes left and the data matrix we are focusing is $2,000 \times 13,182$.

We then explore several tissue types with the largest sample size:

- Breast tumor, which has 1599 samples;
- B cell lymphoma, which has 213 samples;
- Prostate tumor, which has 148 samples;
- Wilms tumor, which has 143 samples.

For each tissue type listed above, we apply Spearman and Pearson on the data belonging to this specific tissue type and obtain the first two dominant sparse eigenvectors. Here we set $R_0 = 100$ for both eigenvectors. For Spearman, we do a normal score transformation Klaassen and Wellner (1997) on the original dataset. We subsequently project the whole dataset to the first two principal components using the obtained eigenvectors. The according 2-dimension visualization is illustrated in Figure 5 and Figure 6.

In Figure 5 and Figure 6 each black point represents a sample and each red point represents a sample belonging to the corresponding tissue type. It can be observed that, in 2D plots learnt by Spearman, the red points are averagely more dense and more close to the border of the sample cluster. The first phenomenon indicates that Spearman has the potential to preserve more common information shared by samples from the same tissue type. The second phenomenon indicates that Spearman has the potential to differentiate samples from different tissue types more efficiently.

## 5.3 Brain Imaging Data

In this section we apply Spearman and Pearson to a brain imaging data: The ADHD 200 dataset (Eloyan et al., 2012). Here 776 subjects' functional scans were collected, where 491 of which are normal persons and 285 of which are diagnosed attention deficit hyperactive disorder (ADHD). The data are normalized and 264 voxels with biological interests are extracted. These voxels broadly cover the major functional regions of the cerebral cortex and cerebellum. We refer to Eloyan et al. (2012) and Power et al. (2011) for details in data preprocessing and voxel definitions. In this manucript we are only interested in the normal persons, leading to a data matrix with 491 rows and 264 columns.

We apply Spearman and Pearson, with $R_0$ set to be 20 in each sparse estimated eigenvector, to the ADHD data, and plot the first principal component against the second, third, and fourth principal components. Figure 7 visualizes the results. Here similar as in Section 5.2, for Spearman, we conduct a normal score transformation on the original data. It can be observed that there are outliers in the principal components calculated by Pearson, which can make the inference based on the principal components very unstable. In contrast, the principal components calculated by Spearman are very concrete and present almost like a bivariate Gaussian distribution.

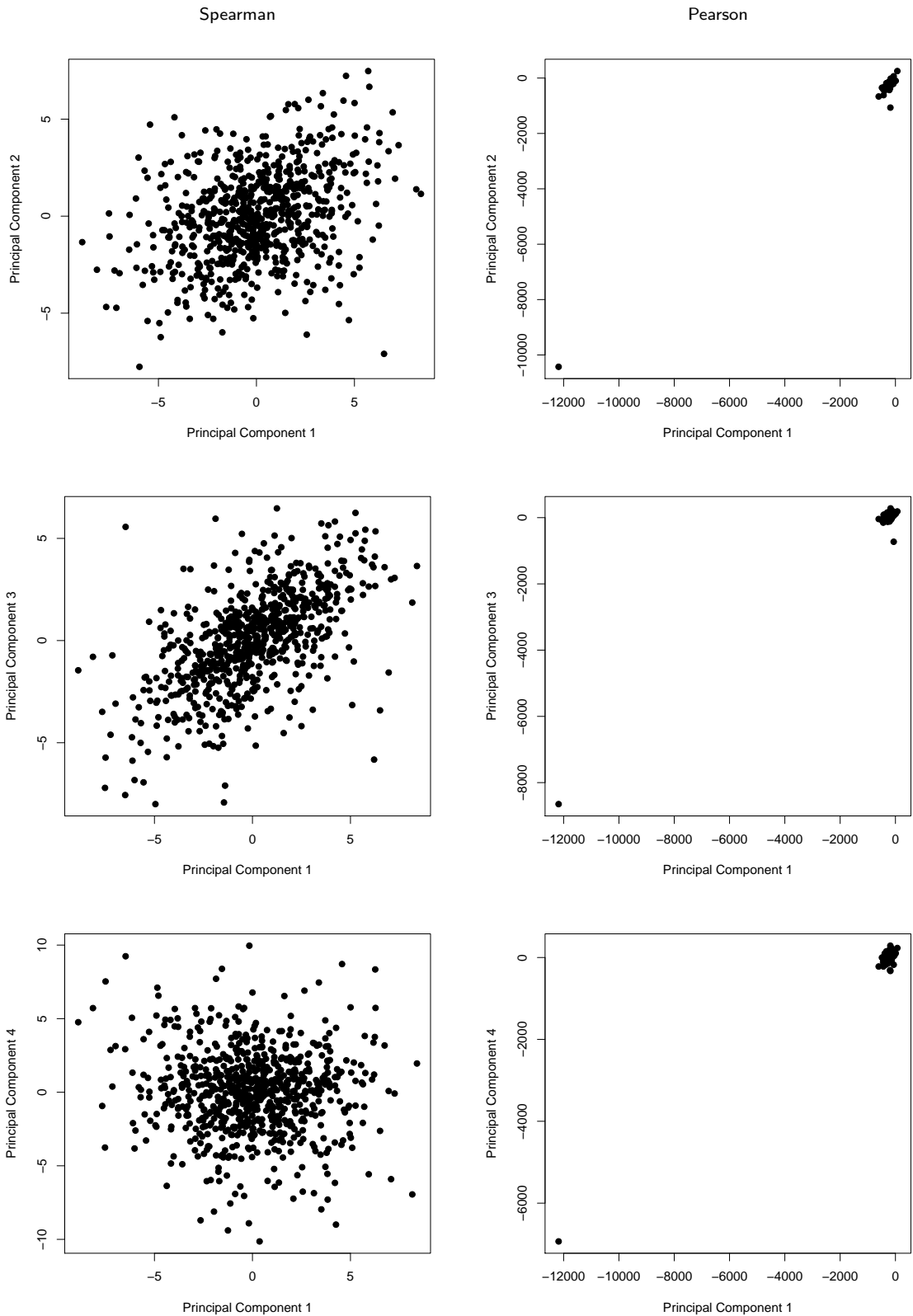

Figure 7: The scatter plots of the first principal component against the second, third, and fourth principal components (from top to bottom) of the brain imaging dataset. Spearman and Pearson are compared ( from left to right)

## 6 Conclusion

In this paper we propose a semiparametric scale-invariant principal component analysis named Copula Component Analysis (COCA). Several contributions we make include: (i) We generalize the Gaussian assumption used in justifying the high dimensional spare PCA to the nonparanormal; (ii) We utilize the rank-based nonparametric correlation coefficient estimator, Spearman's rho, in estimating the latent correlation matrix; (iii) We provide sufficient conditions under which the estimation consistency and feature selection consistency for COCA can be achieved; (iv) We also explore sufficient conditions under which Copula PCA can achieve the same theoretical properties as COCA, and discuss the advantages of COCA over Copula PCA; (v) Careful experimental studies are conducted to confirm that COCA outperforms Copula PCA on both synthetic and real-world datasets.

## A  Supporting Inequalities

**Lemma A.1.** Let $\widehat{\mathbf{R}}$ be the Spearman's rho correlation matrix. Then for any $n \geq \frac{37\pi}{t} + 2$, we have

$$\mathbb{P}\left(|\widehat{\mathbf{R}}_{jk} - \mathbf{\Sigma}^0_{jk}| > t\right) \leq 2\exp\left(-\frac{nt^2}{16\pi^2}\right).$$

*Proof.* Using the notation and proof of Theorem 4.1 in Liu et al. (2012), we have whenever $n \geq \frac{9\pi^2}{(1-\alpha)^2 c^2 \log d}$,

$$\mathbb{P}\left(|\widehat{\mathbf{R}}_{jk} - \mathbb{E}\widehat{\mathbf{R}}_{jk}| > \frac{2c}{\pi}\sqrt{\frac{\log d}{n}}\right) \leq 2\exp\left(-\frac{2\alpha^2 c^2 \log d}{27\pi^2}\right), \tag{A.1}$$

and whenever $t \geq \frac{6\pi}{t} + 2$

$$\mathbb{P}\left(|\widehat{\mathbf{R}}_{jk} - \mathbf{\Sigma}^0_{jk}| > t\right) \leq \mathbb{P}\left(|\widehat{\mathbf{R}}_{jk} - \mathbb{E}\widehat{\mathbf{R}}_{jk}| > \frac{2t}{\pi}\right). \tag{A.2}$$

In Equation (A.1), letting $t = c\sqrt{\frac{\log d}{n}}$ and $\alpha = 3\sqrt{6}/8$ and applying (A.2), we have whenever $n \geq \frac{37\pi}{t}$,

$$\mathbb{P}\left(|\widehat{\mathbf{R}}_{jk} - \mathbf{\Sigma}^0_{jk}| > t\right) \leq 2\exp\left(-\frac{nt^2}{16\pi^2}\right).$$

This completes the proof. □

**Theorem A.2** (Bernstein Inequality). Let $x_1, \ldots, x_n$ be $n$ independent realizations of a random variable $X$ with $\mathbb{E}X = 0$. Suppose that for some positive constant $K$, we have

$$\mathbb{E}|X^m| \leq \frac{m!}{2}K^{m-2}, \quad m = 2, 3, \ldots.$$

Then for all $0 < t \leq \frac{1-2C}{2KC}$,

$$\mathbb{P}\left(\frac{1}{n}\sum_{i=1}^n x_i \geq t\right) \leq \exp(-nCt^2),$$

where $C$ is a generic constant not scaled with $(n, d)$.

*Proof.* Using Lemma 5.7 of Van De Geer (2000), we have for all $a > 0$

$$\mathbb{P}(\sum x_i \geq \sqrt{n}a) \leq \exp\left(-\frac{a^2}{2(aKn^{-1/2} + 1)}\right).$$

Letting $t = a/\sqrt{n}$, we have

$$\mathbb{P}\left(\frac{1}{n}\sum x_i \geq t\right) \leq \exp\left(-\frac{nt^2}{2(Kt + 1)}\right).$$

If $t \leq \frac{1/C-2}{2K}$, we further have $2(Kt + 1) \leq 1/C$, implying that

$$\mathbb{P}\left(\frac{1}{n}\sum x_i \geq t\right) \leq \exp\left(-nCt^2\right).$$

□

## B  Main Proofs

### B.1  Proof of Lemma 3.1

*Proof.* Equation (3.6) holds if and only if

$$\frac{\|\boldsymbol{v}_{A_{k+1}}\|_2}{\|\boldsymbol{v}_{A_k}\|_2} \leq \frac{\|\boldsymbol{v}_{A_{k+1}}\|_q}{\|\boldsymbol{v}_{A_k}\|_q}.$$

This is equivalent to proving that

$$\left(1 + \frac{v_{k+1}^2}{\sum_{j=1}^k v_j^2}\right)^{1/2} \leq \left(1 + \frac{v_{k+1}^q}{\sum_{j=1}^k v_j^q}\right)^{1/q}. \quad (B.1)$$

If $v_{k+1} = 0$, it is easy to see that Equation (B.1) holds. If not, denoting by $m_j = \frac{v_j}{v_{k+1}} \geq 1$, to prove that Equation (B.1) holds is equivalent to proving that

$$\left(1 + \frac{1}{\sum_{j=1}^k m_j^2}\right)^q \leq \left(1 + \frac{1}{\sum_{j=1}^k m_j^q}\right)^2.$$

Realizing that for any $x \in \mathbb{R}^+ \cap \{0\}$ and $0 < \alpha \leq 1$, $x^\alpha \leq 1 + \alpha(x - 1)$, we have

$$\left(1 + \frac{1}{\sum_{j=1}^k m_j^2}\right)^q \leq 1 + q \cdot \frac{1}{\sum_{j=1}^k m_j^2} \leq 1 + 2 \cdot \frac{1}{\sum_{j=1}^k m_j^q} \leq \left(1 + \frac{1}{\sum_{j=1}^k m_j^q}\right)^2.$$

This completes the proof.
□

### B.2  Proof of Lemma 3.2

*Proof.* By using Lemma A.1, we have

$$\mathbb{P}(|\widehat{\mathbf{R}}_{jk} - \boldsymbol{\Sigma}_{jk}^0| > t) \leq 2\exp\left(-\frac{nt^2}{16\pi^2}\right).$$

Because $\boldsymbol{\Sigma}^0$ is feasible to Equation (3.8), $\widetilde{\mathbf{R}}$ must satisfy that: $\|\widehat{\mathbf{R}} - \widetilde{\mathbf{R}}\|_{\max} \leq \|\widehat{\mathbf{R}} - \boldsymbol{\Sigma}^0\|_{\max}$. Using the triangular inequality, we then have

$$\mathbb{P}(|\widetilde{\mathbf{R}}_{jk} - \boldsymbol{\Sigma}^0_{jk}| \geq t)\mathbb{P}(|\widetilde{\mathbf{R}}_{jk} - \widehat{\mathbf{R}}_{jk}| + |\widehat{\mathbf{R}}_{jk} - \boldsymbol{\Sigma}^0_{jk}| \geq t) \leq \mathbb{P}(\|\widetilde{\mathbf{R}} - \widehat{\mathbf{R}}\|_{\max} + \|\widehat{\mathbf{R}} - \boldsymbol{\Sigma}^0\|_{\max} \geq t)$$
$$\leq \mathbb{P}(\|\widehat{\mathbf{R}} - \boldsymbol{\Sigma}^0\|_{\max} \geq t/2) \leq d^2 \exp\left(-\frac{nt^2}{64\pi^2}\right) \leq 2\exp\left(\frac{2\log d}{\log 2} - \frac{nt^2}{64\pi^2}\right).$$

Using the fact that $t \geq 16\pi\sqrt{\frac{\log d}{n \log 2}}$, we have the result. $\square$

## B.3 Proof of Lemma 4.2

*Proof.* Because $g_j^2 \in TF(K)$, where $K$ is a constant not scaled with $(n, d)$, we have that $X_j$'s moments are controlled by $K$ for $j = 1, \ldots, d$. Therefore, $\mu_j$ and $\sigma_j$ are not scaled with $(n, d)$. Accordingly, we can assume that $\boldsymbol{\mu} = \mathbf{0}$ and $\operatorname{diag}(\boldsymbol{\Sigma}) = \mathbf{1}$ without loss of generality. Let $\boldsymbol{X} = (X_1, \ldots, X_d)^T \sim MNPN_d(\boldsymbol{\mu}, \boldsymbol{\Sigma}, f)$. To prove that Equation (4.4) and Equation (4.5) hold, the key is to prove that the high order moments of each $X_j$ and $X_j^2$ will not grow very fast.

Generally, define $Z := f_j(X_j) \sim N(0, 1)$. We have $\forall \ m \in \mathbb{Z}^+$, because $g_j^2 \in \mathrm{TF}(K)$ for some constant $K$, by definition,

$$\mathbb{E}|X_j^2|^m = \mathbb{E}|g_j(Z)^2|^m \leq \frac{m!}{2}K^m.$$

Moreover, we have $\mathbb{E}(X_j)^m$ can be bounded by a similar term. More specifically, if $m$ is even, we have

$$\mathbb{E}|X_j|^m = \mathbb{E}|X_j^2|^{m/2} \leq \frac{(m/2)!}{2}K^{m/2} < \frac{m!}{2}K^m;$$

If $m$ is odd, we have

$$\mathbb{E}|X_j|^m \leq 1 + \mathbb{E}|X_j|^m I(|X_j| \geq 1) \leq 1 + \mathbb{E}|X_j|^{m+1} \leq 1 + \frac{\left(\frac{m+1}{2}\right)!}{2}K^{\frac{m+1}{2}} < \frac{m!}{2}(2K+2)^m.$$

Therefore, realizing that $\widehat{\mu}_j = \frac{1}{n}\sum_{i=1}^n x_{ij}$, $\widehat{\sigma}_j^2 = \frac{1}{n}\sum_{i=1}^n x_{ij}^2$, and $\mathbb{E}X_j^m \leq \frac{m!}{2}(2K+2)^m$ and $\mathbb{E}(X_j^2)^m \leq \frac{m!}{2}K^m$, we can apply the Bernstein inequality (shown in Theorem A.2) to obtain a concentration inequality for $\widehat{\mu}_j$ and $\widehat{\sigma}_j^2$. In particular, we have

$$\mathbb{P}(|\widehat{\mu}_j - \mu_j| > t) \leq 2\exp(-c_2 n t^2) \text{ and } \mathbb{P}(|\widehat{\sigma}_j^2 - \sigma_j^2| > t) \leq 2\exp(-c_3 n t^2)$$

where $c_2$ and $c_3$ only depend on $K$. We further have

$$\mathbb{P}(|\widehat{\sigma}_j - \sigma_j| > t) \leq \mathbb{P}(|\widehat{\sigma}_j^2 - \sigma_j^2| > c_0 t) \leq 2\exp(-c_4 n t^2),$$

where $c_0$ is a generic constant and $c_4 = c_3 \cdot c_0^2$ only depends on the choice the $K$. To finalize the proof, we need to show that combining $\widehat{R}$ with $\{\widehat{\sigma}_1, \ldots, \widehat{\sigma}_d\}$ will not hurt the rate. To show this, suppose that

$$\mathbb{P}\left(|\widehat{\sigma}_j - \sigma_j| > t\right) \leq \eta_1(n, t) \text{ and } \mathbb{P}\left(\left|\widehat{\mathbf{R}}_{jk} - \boldsymbol{\Sigma}^0_{jk}\right| > t\right) \leq \eta_2(n, t).$$

Letting $\sigma^2_{\max} := \max_j(\sigma^2_j)$ be controlled by $K/2$, we have

$$\begin{aligned}
\mathbb{P}\left(\left|\widehat{\mathbf{S}}_{jk} - \mathbf{\Sigma}_{jk}\right| > \epsilon\right) &\leq \mathbb{P}\left(\left|(\widehat{\sigma}_j\widehat{\sigma}_k - \sigma_j\sigma_k)\widehat{\mathbf{R}}_{jk}\right| > \frac{\epsilon}{2}\right) + \mathbb{P}\left(\left|\sigma_j\sigma_k\left(\widehat{\mathbf{R}}_{jk} - \mathbf{\Sigma}^0_{jk}\right)\right| > \frac{\epsilon}{2}\right) \\
&\leq \mathbb{P}\left(\left|\widehat{\sigma}_j\widehat{\sigma}_k - \sigma_j\sigma_k\right| > \frac{\epsilon}{2}\right) + \mathbb{P}\left(\left|\widehat{\mathbf{R}}_{jk} - \mathbf{\Sigma}^0_{jk}\right| > \frac{\epsilon}{2\sigma^2_{\max}}\right) \\
&\leq \mathbb{P}\left(\left|\widehat{\sigma}_j - \sigma_j\right| > \sqrt{\frac{\epsilon}{6}}\right) + \mathbb{P}\left(\left|\widehat{\sigma}_k - \sigma_k\right| > \sqrt{\frac{\epsilon}{6}}\right) + \mathbb{P}\left(\left|\widehat{\sigma}_k - \sigma_k\right| > \frac{\epsilon/6}{\sigma_{max}}\right) \\
&\quad + \mathbb{P}\left(\left|\widehat{\sigma}_j - \sigma_j\right| > \frac{\epsilon}{6\sigma_{max}}\right) + \eta_2\left(n, \frac{\epsilon}{2\sigma^2_{max}}\right) \\
&\leq 2\eta_1\left(n, \sqrt{\frac{\epsilon}{6}}\right) + 2\eta_1\left(n, \frac{\epsilon}{6\sigma_{max}}\right) + \eta_2\left(n, \frac{\epsilon}{2\sigma^2_{max}}\right).
\end{aligned}$$

By using Lemma A.1, we have $\mathbb{P}(|\widehat{\mathbf{R}}_{jk} - \mathbf{\Sigma}^0_{jk}| > t) \leq 2\exp\left(-\frac{nt^2}{16\pi^2}\right)$. It means that $\eta_1$ and $\eta_2$ are both of parametric exponential decay rate. This completes the proof. □

### B.4 Proof of Theorem 4.4

*Proof.* For $\mathcal{M}^0(q, R_q, \mathbf{\Sigma}^0, f^0)$ with $0 \leq q \leq 1$, we define

$$\epsilon := \sin \angle(\boldsymbol{\theta}_1, \widetilde{\boldsymbol{\theta}}_1) \quad \text{and} \quad \mathbf{\Sigma}^0 = \lambda_1 \boldsymbol{\theta}_1 \boldsymbol{\theta}_1^T + \mathbf{\Psi}_0, \tag{B.2}$$

where $\mathbf{\Psi}_0 = \sum_{j=2}^d \lambda_j \boldsymbol{\theta}_j \boldsymbol{\theta}_j^T$ is perpendicular to $\boldsymbol{\theta}_1$. For all $\boldsymbol{\theta} \in \mathbb{S}^{d-1}$, we have

$$\langle \mathbf{\Sigma}^0, \boldsymbol{\theta}_1 \boldsymbol{\theta}_1^T - \boldsymbol{\theta}\boldsymbol{\theta}^T \rangle = \langle \mathbf{\Sigma}^0, \boldsymbol{\theta}_1\boldsymbol{\theta}_1^T \rangle - \langle \lambda_1 \boldsymbol{\theta}_1\boldsymbol{\theta}_1^T + \mathbf{\Psi}_0, \boldsymbol{\theta}\boldsymbol{\theta}^T \rangle = \lambda_1 - \lambda_1 \langle \boldsymbol{\theta}_1, \boldsymbol{\theta}\rangle^2 - \langle \mathbf{\Psi}_0, \boldsymbol{\theta}\boldsymbol{\theta}^T\rangle, \tag{B.3}$$

$$\langle \mathbf{\Psi}_0, \boldsymbol{\theta}\boldsymbol{\theta}^T \rangle = \boldsymbol{\theta}^T \mathbf{\Psi}_0 \boldsymbol{\theta} = \boldsymbol{\theta}^T(\mathbf{I}_d - \boldsymbol{\theta}_1\boldsymbol{\theta}_1^T)\mathbf{\Sigma}^0(\mathbf{I}_d - \boldsymbol{\theta}_1\boldsymbol{\theta}_1^T)\boldsymbol{\theta} \leq \lambda_2 \|(\mathbf{I}_d - \boldsymbol{\theta}_1\boldsymbol{\theta}_1^T)\boldsymbol{\theta}\|_2^2 = \lambda_2 - \lambda_2 \langle \boldsymbol{\theta}_1, \boldsymbol{\theta}\rangle^2. \tag{B.4}$$

Moreover, by definition, we have

$$\sin^2 \angle(\boldsymbol{\theta}_1, \boldsymbol{\theta}) = 1 - (\boldsymbol{\theta}_1^T \boldsymbol{\theta})^2 = 1 - \langle \boldsymbol{\theta}_1, \boldsymbol{\theta}\rangle^2. \tag{B.5}$$

Combining Equation (B.3) with Equation (B.5), we have

$$\langle \mathbf{\Sigma}^0, \boldsymbol{\theta}_1\boldsymbol{\theta}_1^T - \boldsymbol{\theta}\boldsymbol{\theta}^T \rangle \geq (\lambda_1 - \lambda_2)\sin^2 \angle(\boldsymbol{\theta}_1, \boldsymbol{\theta}).$$

Therefore, letting $\widetilde{\boldsymbol{\theta}}_1$ be the minimizer to Equation (3.2), we have

$$\begin{aligned}
\epsilon^2 &\leq \frac{1}{\lambda_1 - \lambda_2}\left\langle \mathbf{\Sigma}^0, \boldsymbol{\theta}_1\boldsymbol{\theta}_1^T - \widetilde{\boldsymbol{\theta}}_1\widetilde{\boldsymbol{\theta}}_1^T \right\rangle \leq \frac{1}{\lambda_1 - \lambda_2}\left(\left\langle \mathbf{\Sigma}^0 - \widehat{\mathbf{R}}, \boldsymbol{\theta}_1\boldsymbol{\theta}_1^T - \widetilde{\boldsymbol{\theta}}_1\widetilde{\boldsymbol{\theta}}_1^T\right\rangle + \left\langle \widehat{\mathbf{R}}, \boldsymbol{\theta}_1\boldsymbol{\theta}_1^T - \widetilde{\boldsymbol{\theta}}_1\widetilde{\boldsymbol{\theta}}_1^T \right\rangle\right) \\
&\leq \frac{1}{\lambda_1 - \lambda_2}\left\langle \mathbf{\Sigma}^0 - \widehat{\mathbf{R}}, \boldsymbol{\theta}_1\boldsymbol{\theta}_1^T - \widetilde{\boldsymbol{\theta}}_1\widetilde{\boldsymbol{\theta}}_1^T \right\rangle.
\end{aligned} \tag{B.6}$$

The last inequality holds because

$$\left\langle \widehat{\mathbf{R}}, \boldsymbol{\theta}_1\boldsymbol{\theta}_1^T - \widetilde{\boldsymbol{\theta}}_1\widetilde{\boldsymbol{\theta}}_1^T \right\rangle = \boldsymbol{\theta}_1^T \widehat{\mathbf{R}} \boldsymbol{\theta}_1 - \widetilde{\boldsymbol{\theta}}_1^T \widehat{\mathbf{R}} \widetilde{\boldsymbol{\theta}}_1 \leq 0.$$

Therefore, using Equation (B.6),

$$\begin{aligned}
\epsilon^2 &\leq \frac{1}{\lambda_1 - \lambda_2}\left\langle \mathbf{\Sigma}^0 - \widehat{\mathbf{R}}, \boldsymbol{\theta}_1\boldsymbol{\theta}_1^T - \widetilde{\boldsymbol{\theta}}_1\widetilde{\boldsymbol{\theta}}_1^T \right\rangle = \frac{1}{\lambda_1 - \lambda_2}\left\langle \text{vec}(\mathbf{\Sigma}^0 - \widehat{\mathbf{R}}), \text{vec}(\boldsymbol{\theta}_1\boldsymbol{\theta}_1^T - \widetilde{\boldsymbol{\theta}}_1\widetilde{\boldsymbol{\theta}}_1^T)\right\rangle \\
&\leq \frac{1}{\lambda_1 - \lambda_2}\|\text{vec}(\widehat{\mathbf{R}} - \mathbf{\Sigma}^0)\|_\infty \cdot \|\text{vec}(\boldsymbol{\theta}_1\boldsymbol{\theta}_1^T - \widetilde{\boldsymbol{\theta}}_1\widetilde{\boldsymbol{\theta}}_1^T)\|_1,
\end{aligned} \tag{B.7}$$

where the last inequality is by using Hölder Inequality.

**When** $q = 1$, we have

$$\|\operatorname{vec}(\boldsymbol{\theta}_1\boldsymbol{\theta}_1^T - \widetilde{\boldsymbol{\theta}}_1\widetilde{\boldsymbol{\theta}}_1^T)\|_1 \leq \|\operatorname{vec}(\boldsymbol{\theta}_1\boldsymbol{\theta}_1^T)\|_1 + \|\operatorname{vec}(\widetilde{\boldsymbol{\theta}}_1\widetilde{\boldsymbol{\theta}}_1^T)\|_1 = \sum_j\sum_k |\theta_{1j}\theta_{1k}| + \sum_j\sum_k |\widetilde{\theta}_{1j}\widetilde{\theta}_{1k}|$$
$$= \|\boldsymbol{\theta}_1\|_1^2 + \|\widetilde{\boldsymbol{\theta}}_1\|_1^2 \leq 2R_1^2. \qquad (B.8)$$

The last inequality holds because both $\widetilde{\boldsymbol{\theta}}_1$ and $\boldsymbol{\theta}_1$ belong to $\mathbb{B}_1(R_1)$. Therefore,

$$\epsilon^2 \leq 2R_1^2 \cdot \frac{\|\operatorname{vec}(\widehat{\mathbf{R}} - \boldsymbol{\Sigma}^0)\|_\infty}{\lambda_1 - \lambda_2}. \qquad (B.9)$$

**When** $0 < q < 1$, letting $\boldsymbol{\vartheta} = \operatorname{vec}(\boldsymbol{\theta}_1\boldsymbol{\theta}_1^T - \widetilde{\boldsymbol{\theta}}_1\widetilde{\boldsymbol{\theta}}_1^T)$, we have

$$\|\boldsymbol{\vartheta}\|_q^q \leq \|\operatorname{vec}(\boldsymbol{\theta}_1\boldsymbol{\theta}_1^T)\|_q^q + \|\operatorname{vec}(\widetilde{\boldsymbol{\theta}}_1\widetilde{\boldsymbol{\theta}}_1^T)\|_q^q = \sum_j\sum_k (\theta_{1j}\theta_{1k})^q + \sum_j\sum_k (\widetilde{\theta}_{1j}\widetilde{\theta}_{1k})^q = \|\boldsymbol{\theta}_1\|_q^q + \|\widetilde{\boldsymbol{\theta}}_1\|_q^q \leq 2R_q^2.$$

Therefore, denoting by $S_\vartheta := \{j, |\vartheta_j| > \tau\}$ for some $\tau$, we have

$$\|\boldsymbol{\vartheta}\|_1 = \|\boldsymbol{\vartheta}_{S_\vartheta}\|_1 + \sum_{j \notin S_\vartheta} |\vartheta_j| \leq \sqrt{\operatorname{card}(S_\vartheta)}\|\boldsymbol{\vartheta}\|_2 + \tau\sum_{j \notin S_\vartheta}\frac{|\vartheta_j|}{\tau} \leq \sqrt{\operatorname{card}(S_\vartheta)}\|\boldsymbol{\vartheta}\|_2 + \tau\sum_{j \notin S_\vartheta}\left(\frac{|\vartheta_j|}{\tau}\right)^q$$
$$\leq \sqrt{2}R_q\tau^{-q/2}\|\boldsymbol{\vartheta}\|_2 + 2R_q^2\tau^{1-q}. \qquad (B.10)$$

The last inequality holds because

$$\operatorname{card}(S_\vartheta) \cdot \tau^q \leq \sum_{j \in S_\vartheta} |\vartheta_j|^q \leq \|\boldsymbol{\vartheta}\|_q^q \leq 2R_q^2 \quad \text{and} \quad \sum_{j \notin S_\vartheta}|\vartheta_j|^q \leq \|\boldsymbol{\vartheta}\|_q^q \leq 2R_q^2.$$

Therefore, letting $\tau = \frac{\|\operatorname{vec}(\widehat{\mathbf{R}} - \boldsymbol{\Sigma}^0)\|_\infty}{\lambda_1 - \lambda_2}$ and realizing that $\|\boldsymbol{\vartheta}\|_2^2 = \|\operatorname{vec}(\boldsymbol{\theta}_1\boldsymbol{\theta}_1^T - \widetilde{\boldsymbol{\theta}}_1\widetilde{\boldsymbol{\theta}}_1^T)\|_2^2 = 2(1 - (\boldsymbol{\theta}_1^T\widetilde{\boldsymbol{\theta}}_1)^2) = 2\epsilon^2$, combining Equation (B.7) with (B.10), we have

$$\epsilon^2 \leq 2\tau^{1-q/2}R_q\epsilon + 2\tau^{2-q}R_q^2.$$

Therefore, $\epsilon \leq (1 + \sqrt{3})\tau^{1-q/2}R_q$, in other words,

$$\epsilon^2 \leq (1 + \sqrt{3})^2 R_q^2 \left(\frac{\|\operatorname{vec}(\widehat{\mathbf{R}} - \boldsymbol{\Sigma}^0)\|_\infty}{\lambda_1 - \lambda_2}\right)^{2-q}. \qquad (B.11)$$

**When** $q = 0$, denoting by $\boldsymbol{\vartheta} = \operatorname{vec}(\boldsymbol{\theta}_1\boldsymbol{\theta}_1^T - \widetilde{\boldsymbol{\theta}}_1\widetilde{\boldsymbol{\theta}}_1^T)$,

$$\|\boldsymbol{\vartheta}\|_1 \leq \sqrt{\operatorname{card}(\operatorname{supp}(\boldsymbol{\vartheta}))}\|\boldsymbol{\vartheta}\|_2 \leq \sqrt{2R_0^2} \cdot \sqrt{2\epsilon^2} = 2R_0\epsilon. \qquad (B.12)$$

Therefore, combining Equation (B.7) with Equation (B.12), we have $\epsilon^2 \leq \frac{\|\operatorname{vec}(\widehat{\mathbf{R}} - \boldsymbol{\Sigma}^0)\|_\infty}{\lambda_1 - \lambda_2} \cdot 2R_0\epsilon$, which is equivalent to stating that

$$\epsilon^2 \leq 4R_0^2 \left(\frac{\|\operatorname{vec}(\widehat{\mathbf{R}} - \boldsymbol{\Sigma}^0)\|_\infty}{\lambda_1 - \lambda_2}\right)^2. \qquad (B.13)$$

Combining Equation (B.9), (B.11) and (B.13), we have that for all $0 \leq q \leq 1$:

$$\epsilon^2 \leq \gamma_q R_q^2 \left(\frac{\|\operatorname{vec}(\widehat{\mathbf{R}} - \boldsymbol{\Sigma}^0)\|_\infty}{\lambda_1 - \lambda_2}\right)^{2-q}, \quad \text{where } \gamma_q = 2 \cdot I(q=1) + 4 \cdot I(q=0) + (1+\sqrt{3})^2 \cdot I(0 < q < 1).$$

Then, using Lemma 4.2, we have

$$\begin{aligned}\mathbb{P}(\epsilon^2 \geq t) \leq &\mathbb{P}\left(\frac{\gamma_q R_q^2}{(\lambda_1 - \lambda_2)^{2-q}}\|\operatorname{vec}(\widehat{\mathbf{R}} - \mathbf{\Sigma}^0)\|_\infty^{2-q} \geq t\right) = \mathbb{P}\left(\|\widehat{\mathbf{R}} - \mathbf{\Sigma}^0\|_{\max} \geq \left(\frac{t(\lambda_1 - \lambda_2)^{2-q}}{\gamma_q R_q^2}\right)^{\frac{1}{2-q}}\right)\\ \leq &d^2 \exp\left(-\frac{n}{16\pi^2} \cdot \left(\frac{t(\lambda_1 - \lambda_2)^{2-q}}{\gamma_q R_q^2}\right)^{2/(2-q)}\right),\end{aligned} \qquad (\text{B.14})$$

where in the last inequality the constant $\frac{1}{16\pi^2}$ is derived by using Lemma A.1. Finally, choosing $t = \gamma_q R_q^2 \left(\frac{64\pi^2}{(\lambda_1 - \lambda_2)^2} \cdot \frac{\log d}{n}\right)^{\frac{2-q}{2}}$, we have the result. $\square$

### B.5 Proof of Corollary 4.2

*Proof.* Without loss of generality and for simplicity, we may assume that $\widetilde{\boldsymbol{\theta}}_1^T \boldsymbol{\theta}_1 \geq 0$, because otherwise we can simply do appropriate sign changes in the proof. We first note that

$$\operatorname{card}(\widehat{\Theta}^0) = \operatorname{card}(\Theta^0) = R_0. \qquad (\text{B.15})$$

If $\widehat{\Theta}^0 \neq \Theta^0$, then let $\widehat{\Theta}_d := (\widehat{\Theta}^0/\Theta^0) \cup (\Theta^0/\widehat{\Theta}^0)$. We have

$$\|\widetilde{\boldsymbol{\theta}}_1 - \boldsymbol{\theta}_1\|_2^2 \geq \|(\widetilde{\boldsymbol{\theta}}_1 - \boldsymbol{\theta}_1)_{\widehat{\Theta}_d}\|_2^2 = \sum_{j \in \widehat{\Theta}_d}(\widetilde{\theta}_{1j}^2 + \theta_{1j}^2).$$

The last equality holds because for any $j \in \widehat{\Theta}_d$, either $\widetilde{\theta}_{1j}$ or $\theta_{1j}$ are non-zero. Because of Equation (B.15), if $\widehat{\Theta}_d \neq \emptyset$, there must exist $j \in \widehat{\Theta}_d$ such that $\theta_{1j} \neq 0$. Therefore,

$$\|\widetilde{\boldsymbol{\theta}}_1 - \boldsymbol{\theta}_1\|_2 \geq \min_{j \in \Theta}|\theta_{1j}| \geq \frac{16\sqrt{2}R_0\pi}{\lambda_1 - \lambda_2}\sqrt{\frac{\log d}{n}}.$$

Then we have $\sin^2 \angle(\widetilde{\boldsymbol{\theta}}_1, \boldsymbol{\theta}_1) = 1 - (\widetilde{\boldsymbol{\theta}}_1^T\boldsymbol{\theta}_1)^2 \geq 1 - \widetilde{\boldsymbol{\theta}}_1^T\boldsymbol{\theta}_1 = \frac{\|\boldsymbol{\theta}_1 - \widetilde{\boldsymbol{\theta}}_1\|_2^2}{2}$, implying that $\sin^2 \angle(\widetilde{\boldsymbol{\theta}}_1, \boldsymbol{\theta}_1) \geq \frac{\|\widetilde{\boldsymbol{\theta}}_1 - \boldsymbol{\theta}_1\|_2^2}{2} \geq \frac{256R_0^2\pi^2}{(\lambda_1 - \lambda_2)^2} \cdot \frac{\log d}{n}$. Therefore, employing Theorem 4.4, we have

$$\mathbb{P}(\widehat{\Theta}^0 \neq \Theta^0) \leq \mathbb{P}\left(\sin^2 \angle(\widetilde{\boldsymbol{\theta}}_1, \boldsymbol{\theta}_1) \geq \frac{256R_0^2\pi^2}{(\lambda_1 - \lambda_2)^2} \cdot \frac{\log d}{n}\right) \leq \frac{1}{d^2}.$$

This completes the proof. $\square$

### B.6 Proof of Theorem 4.7

*Proof.* We first prove that $\max_{\boldsymbol{v} \in \mathbb{S}^{d-1} \cap \mathbb{B}_0(R_0 + 2k)}|\boldsymbol{v}^T(\widehat{\mathbf{R}} - \mathbf{\Sigma}^0)|$ can be bounded by $8\pi(R_0 + 2k)\sqrt{\frac{\log d}{n}}$ with large probability. To show that, we have

$$|\boldsymbol{v}^T(\widehat{\mathbf{R}} - \mathbf{\Sigma}^0)\boldsymbol{v}| = \left|\left\langle \widehat{\mathbf{R}} - \mathbf{\Sigma}^0, \boldsymbol{v}\boldsymbol{v}^T\right\rangle\right| = \left|\left\langle \operatorname{vec}(\widehat{\mathbf{R}} - \mathbf{\Sigma}^0), \operatorname{vec}(\boldsymbol{v}\boldsymbol{v}^T)\right\rangle\right| \leq \|\operatorname{vec}\widehat{\mathbf{R}} - \mathbf{\Sigma}^0\|_\infty \cdot \|\operatorname{vec}(\boldsymbol{v}\boldsymbol{v}^T)\|_1$$
$$\leq \|\widehat{\mathbf{R}} - \mathbf{\Sigma}^0\|_{\max} \cdot (R_0 + 2k).$$

Using Lemma 2.3, we have the result. Then replacing $\rho(E, s)$ with $8\pi(R_0 + 2k)\sqrt{\frac{\log d}{n}}$ in Theorem 1 in Yuan and Zhang (2013) and realizing that for any $\boldsymbol{v}_1, \boldsymbol{v}_2 \in \mathbb{S}^{d-1}$

$$\sqrt{1 - |\boldsymbol{v}_1^T\boldsymbol{v}_2|} \leq \sqrt{1 - |\boldsymbol{v}_1\boldsymbol{v}_2|^2} = \sin \angle(\boldsymbol{v}_1, \boldsymbol{v}_2) = \sqrt{1 + |\boldsymbol{v}_1^T\boldsymbol{v}_2|} \cdot \sqrt{1 - |\boldsymbol{v}_1^T\boldsymbol{v}_2|} \leq 2\sqrt{1 - |\boldsymbol{v}_1^T\boldsymbol{v}_2|},$$

we have the result. $\square$